\documentclass[journal,twoside]{IEEEtran}

\usepackage{graphicx}
\usepackage{wrapfig}
\usepackage{colortbl}
\usepackage{float}
\usepackage{overpic}
\usepackage{multirow}
\usepackage{booktabs}
\usepackage{bm}
\usepackage{verbatim}
\usepackage{color}
\usepackage{arydshln}
\usepackage{makecell,rotating}
\usepackage{array}
\usepackage{amsmath}
\usepackage{amssymb}
\usepackage{amsthm}
\usepackage{graphicx}
\usepackage{subfigure}
\usepackage{cite}
\usepackage{color}
\usepackage{epstopdf}
\usepackage{subeqnarray}
\usepackage{cases}
\usepackage{mathrsfs}
\usepackage{multirow}
\usepackage{algorithm}
\usepackage{algorithmicx}
\usepackage{algpseudocode}
\floatname{algorithm}{Algorithm}

\newcommand{\myPara}[1]{\vspace{5pt}\noindent$\bullet$~\textbf{#1} \quad}
\usepackage[dvipsnames]{xcolor}

\usepackage{enumitem}
\setenumerate[1]{itemsep=4pt,partopsep=0pt,parsep=\parskip,topsep=4pt}
\setitemize[1]{itemsep=4pt,partopsep=0pt,parsep=\parskip,topsep=4pt}

\definecolor{citecolor}{RGB}{119,185,0}
\definecolor{citecolor1}{RGB}{66,168,235}
\usepackage{hyperref}
\hypersetup{colorlinks=true,citecolor=citecolor1,linkcolor=BrickRed,urlcolor=Thistle}

\definecolor{mygray}{gray}{.88}
\definecolor{mygrayd}{gray}{.94}

\def\ie{\emph{i.e.}}

\def\etc{\emph{etc}}
\def\etal{{\em et al.~}}

\begin{document}

\title{Learning Visual Affordance Grounding from Demonstration Videos}

\author{Hongchen Luo, Wei Zhai, Jing Zhang, \IEEEmembership{Member, IEEE}, Yang Cao, \IEEEmembership{Member, IEEE}, and Dacheng Tao, \IEEEmembership{Fellow, IEEE}

\thanks{H. Luo, W. Zhai and Y. Cao are with the School of Information Science and Technology, at the University of Science and Technology of China, Anhui, China. (email: \{lhc12, wzhai056\}@mail.ustc.edu.cn, forrest@ustc.edu.cn).}

\thanks{J. Zhang and D. Tao are with the School of Computer Science, in the Faculty of Engineering, at the University of Sydney, Sydney, Australia. (email: \{jing.zhang1,dacheng.tao\}@sydney.edu.au).}

}

\maketitle

\begin{abstract}
Visual affordance grounding aims to segment all possible interaction regions between people and objects from an image/video, which is beneficial for many applications, such as robot grasping and action recognition. However, existing methods mainly rely on the appearance feature of the objects to segment each region of the image, which face the following two problems: (i) there are multiple possible regions in an object that people interact with; and (ii) there are multiple possible human interactions in the same object region. To address these problems, we propose a Hand-aided Affordance Grounding Network (HAG-Net) that leverages the aided clues provided by the position and action of the hand in demonstration videos to eliminate the multiple possibilities and better locate the interaction regions in the object. Specifically, HAG-Net has a dual-branch structure to process the demonstration video and object image. For the video branch, we introduce hand-aided attention to enhance the region around the hand in each video frame and then use the LSTM network to aggregate the action features. For the object branch, we introduce a semantic enhancement module (SEM) to make the network focus on different parts of the object according to the action classes and utilize a distillation loss to align the output features of the object branch with that of the video branch and transfer the knowledge in the video branch to the object branch. Quantitative and qualitative evaluations on two challenging datasets show that our method has achieved state-of-the-art results for affordance grounding. 
The source code will be made available to the public.
\end{abstract}

\begin{IEEEkeywords}
Visual affordance grounding, Learning from demonstrations, Weakly supervised learning, Deep learning.
\end{IEEEkeywords}

\section{Introduction}

\IEEEPARstart{F}{or} an object in the scene, we need to know what it is and what interactions it may have with people, and the possible interactions are the affordance of the object. The concept of affordance was introduced by the ecological psychologist Gibson \cite{theaff} in 1979 to describe the potential ``action possibilities'' that the environment can provide. The ``action possibilities'' reflect all possible interactions between the object and the agent. In particular, perceiving object affordance is a valuable capability and has a wide range of applications in action recognition, robot grasping, autonomous driving, scene understanding, \etc. \cite{zhang2020empowering,DBLP:journals/corr/abs-1807-06775,DBLP:journals/cviu/KjellstromRK11,qi2017predicting,vu2014predicting,yamanobe2017brief,liu2019auto}.

\begin{figure}[t]
	\centering
		\includegraphics[width=1\linewidth]{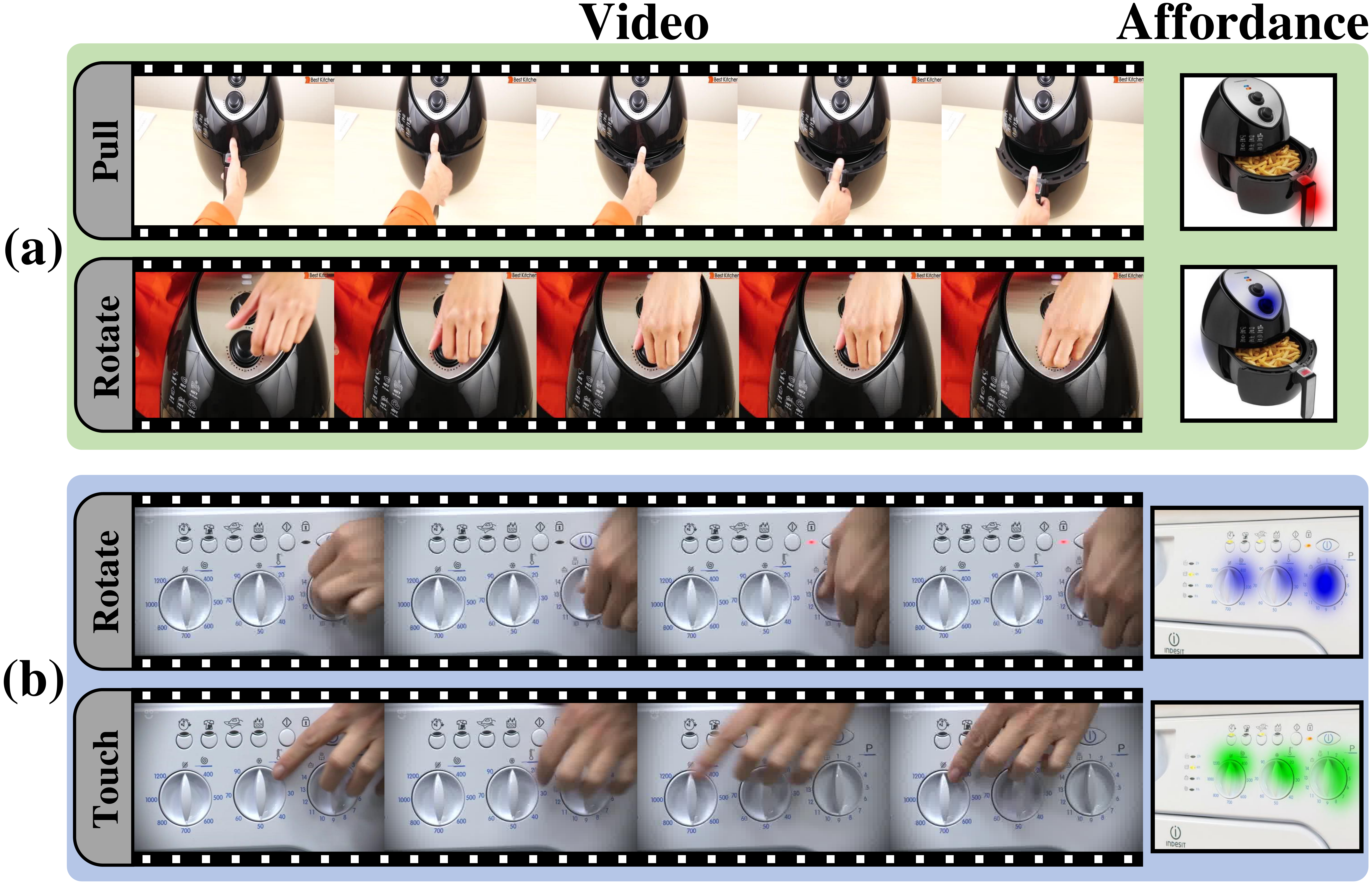}
	\caption{\textbf{The challenges of affordance grounding.} We show some examples of affordance with multiple possibilities. (a) Different interactions occur at different locations of the same object. (b) Different interactions occur at a similar position of the same object.}
	\label{FIG:1}
\end{figure}

This paper focuses on the affordance grounding task, \ie, given an action label and an object image,  our goal is to generate the heatmap where the action happens. However, in Gibson's definition of affordance, ``it implies the complementarity of the animal and the environment'' \cite{theaff}. The multiple potential complementarities between the animal and the environment, leading to the possibility that affordance has multiple possibilities: (i) there are multiple possible regions in an object where people interact. As shown in Fig \ref{FIG:1} (a), the same object may imply two different interactions, ``pull'' and ``rotate'', and the interaction occurs at different locations; and (ii) there are multiple possible human interactions in the same region. As shown in Fig. \ref{FIG:1} (b), two different interactions, ``rotate'' and ``touch'', can happen at the same position on the same object. Existing methods \cite{DBLP:KoppulaS14,DBLP:conf/iros/NguyenKCT17,8460902,DBLP:conf/cvpr/SawatzkySG17} typically establish the mapping relationship between apparent features to affordance labels, which are not skilled at capturing the affordance-related context and perceiving the affordance multiple possibilities.

At the same time, the study \cite{heft1989affordances} found that, in a particular scene, the affordance of the object is uniquely determined by the intention of the human action. For example, in a demonstration video, human hands and objects interact frequently, and human action intention can be inferred from hand position and action, providing an essential clue to solve the above two problems. For the first question, we can infer the region where the person interacts with the object by the hand's position, such as ``touch'' and ``pull'' in Fig. \ref{FIG:1} (a). For the second question, we eliminate the ambiguity of affordance by inferring human intention through human hand actions, such as (b) in Fig. \ref{FIG:1}, according to the hand action in the video, we can infer whether the object's affordance is ``touch'' or ``rotate''. To this end, we both consider the affordance clues provided by the position and the action of the hand and solve multiple possibilities of affordance grounding by establishing a mapping relationship between hand action intentions and object parts.

This paper presents a Hand-aided Affordance Grounding Network (HAG-Net) to learn the affordance of objects from demonstration videos. We use the clues provided by the position and the action of the hand to solve the two problems in the affordance grounding. The pipeline of our method is shown in Fig. \ref{FIG:2}. To better capture the affordance cues from the demonstration videos, we propose a hand-related selection network to select the keyframes which primarily reflect the interaction between the hand and the object.  During learning the object's affordance from the demonstration videos, we emphasize the features around the hand to provide valuable context to learn the region where the person interacts with the object. Instead of simply feeding the hand features directly into the network, we consider the occluded features and the objects surrounding the hand. Since the region of human-object interaction is related to the category of action, we design a semantic enhancement module (SEM) that considers both the action category and appearance features to jointly adjust the feature of the object image so that the network can better focus on the feature regions related to the action category. Finally, we consider both the position and the action of the hand to better transfer the knowledge of the video branch to the object branch. The contributions of this paper are as follows:

\begin{enumerate}

\item 
We propose a Hand-aided Affordance Grounding Network (HAG-Net) to use the clues provided by the hands in the demonstration videos. We leverage both the position of the hand and the action to eliminate the multiple possibilities in the affordance grounding task. 

\item 
We introduce the Semantic Enhancement Module (SEM), which utilizes the action category and appearance features to jointly modulate the feature of the object image to locate the relevant region of the interaction more effectively.

\item 
Experimental results on the two most challenging affordance grounding datasets have demonstrated the superiority of the proposed model against the state-of-the-arts.

\end{enumerate}

\section{Related Work}
\subsection{Affordance Grounding}
Affordance grounding is to detect the regions of the object where interactions may occur. Early works \cite{DBLP:KoppulaS14,DBLP:conf/iros/NguyenKCT17,8460902,2019Object} are mainly on image-based supervised segmentation tasks, outputting the affordance label of each pixel for a given image. However, these tasks rely on pixel-level labels, and these methods can not learn how humans interact with objects.

Another part of the early works represents affordance as a pose in which humans interact with objects. Yao \etal \cite{DBLP:conf/iccv/YaoMF13} used a clustering method to find that all possible object functionalities and represented it in the form of a pose in which human interaction with musical instruments. Wang \etal \cite{Wang_affordanceCVPR2017} used Variational-Auto Encoders (VAE) to construct their model to predict affordance poses. Li \etal \cite{3d-affordance} proposed a 3D generative model to predict physically plausible and physically feasible human poses in a given 3D scene. The difference from the above methods is that we focus on the possible interactions of the various parts of the object instead of the human pose. 

A large number of works solve the affordance grounding by learning the interactions between humans and objects in a video. Koppula \etal \cite{DBLP:KoppulaS14} proposed a generative model to ground the affordance into the form of the object's spatial position and time trajectory. Fang \etal \cite{demo2vec2018cvpr} used the feature embedding of the demonstration video to predict the interaction region of the target object and proposed the OPRA (Online Product Review dataset for Affordance) dataset. Nagarajan \etal \cite{interaction-hotspots} learned the interaction between humans and objects by observing videos. Unlike the above methods, we explicitly infer human action intentions from the cues provided by the position and action of a human hand to specify the uniqueness of affordance in a given scene, thus eliminating ambiguity due to the multiple possibilities of affordance.

\begin{figure}[t]
	\centering
		\includegraphics[width=1\linewidth]{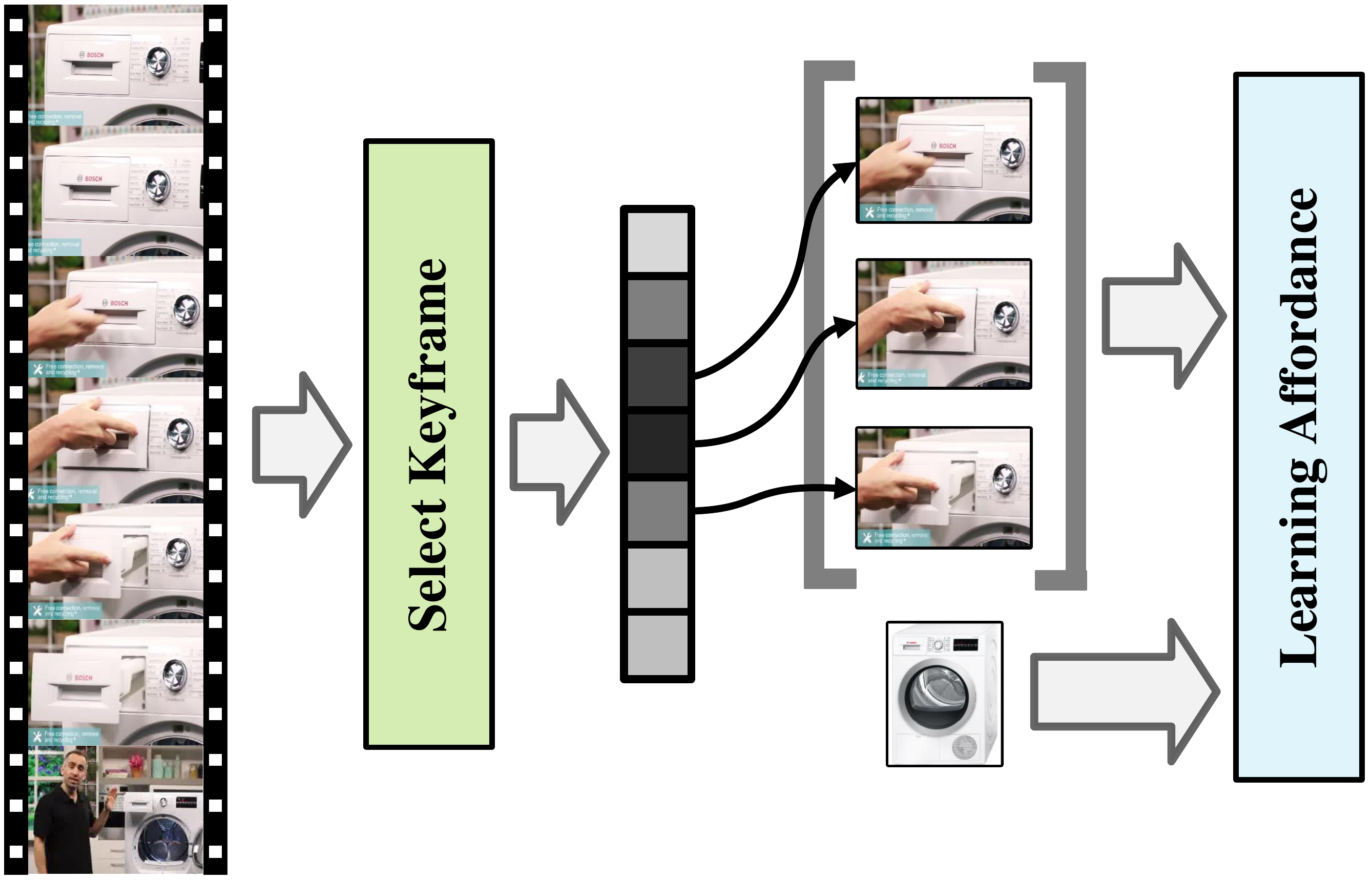}
	\caption{\textbf{The pipeline of our method.} Our method first trains a hand-related selection network to select keyframes and then sends the pre-processed video frames to the Hand-aided Affordance Grounding Network (HAG-Net), trained using only action classes as supervisory signals.}
	\label{FIG:2}
\end{figure}

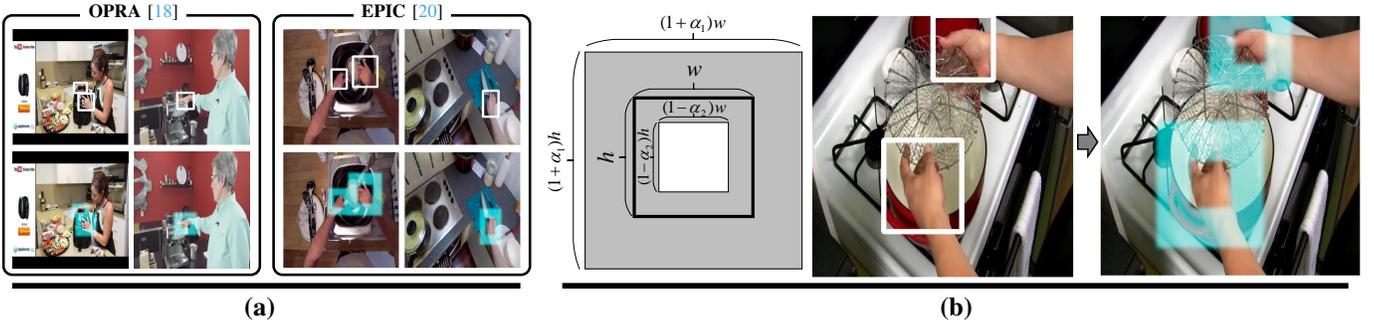
\begin{figure*}[t]
	\centering
		\begin{overpic}[width=1\linewidth]{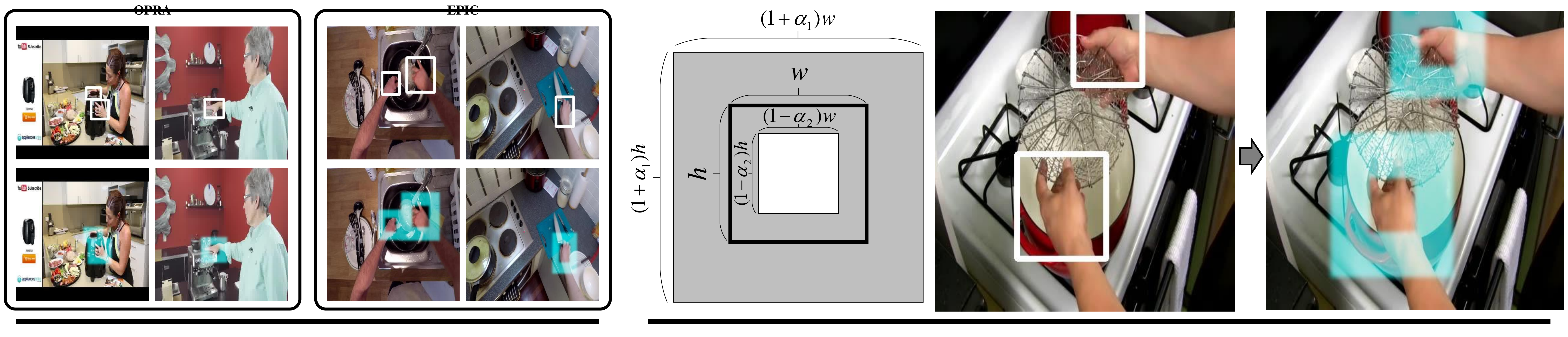}
		    \put(18, -1.2){\textbf{(a)}}
		    \put(69, -1.2){\textbf{(b)}}
		    \put(6, 20.7){\colorbox{white}{\scriptsize\textbf{OPRA} \cite{demo2vec2018cvpr}}}
            \put(26, 20.7){\colorbox{white}{\scriptsize\textbf{EPIC} \cite{Damen2018EPICKITCHENS}}}
		\end{overpic}
	\caption{\textbf{Hand-aided mask}. (a) Some hand detection results, We use the trained YOLOv3 pruning model \cite{yolov3,Liu2017learning}  to detect the position of the hand on each frame of the two datasets of OPRA \cite{demo2vec2018cvpr} and EPIC \cite{Damen2018EPICKITCHENS}. (b) The shadow regions are the surrounding regions of the hand enhanced by hand-aided attention. We extend the bounding box of the hand to the surroundings and remove the regions occluded by the hand. The blue mask on the right side is the hand mask ($Mask_t$) that we input into the network.}
	\label{FIG:3}
\end{figure*}

\subsection{Hand and Affordance}
There is a large number of researches in the field of hand-based egocentric action recognition. Hand detection, segmentation, and tracking technologies \cite{DBLP:conf/cvpr/LiK13,DBLP:conf/cvpr/Betancourt14,DBLP:journals/cviu/BetancourtMBMRR17} continue to develop and their research results are applied to model actions and activities \cite{DBLP:conf/iccv/BambachLCY15,9060114}. Kjellstrom \etal \cite{DBLP:journals/cviu/KjellstromRK11} learned the affordance of objects from the human demonstration and extracted hand position, orientation, and articulated pose for embedding action features. Stark \etal \cite{DBLP:conf/icvs/StarkLZWS08} obtained the affordance clues from the interaction between the human hand and the objects in the trainset and then determined the functions of the objects according to the affordance cue features. Sun \etal \cite{DBLP:journals/ras/0004RL14} significantly improved the detection accuracy of the interactive objects by learning hand movement trajectories and statistical knowledge in training data. Song \etal \cite{DBLP:conf/icra/SongKOPABK13} predicted the human intention by observing the human/object interaction. Pieropan \etal \cite{DBLP:conf/icra/PieropanEK13} directly represent objects as their interactions with human hands for modeling human activity. In this paper, we use the clues provided by the hand to learn the affordance of the objects.

\subsection{Learning from Demonstrations}
By learning how humans interact with objects through demonstrations, the robot can perceive the affordance of the objects. Furthermore, when faces with objects in different environments, the robot can mimic human actions to interact with the objects through the knowledge learned from human demonstrations. In recent years, there has been a great deal of works on learning from demonstrations \cite{schulman2016learning,chu2016learning,aleotti2011part,zha2021contrastively,demo2vec2018cvpr}.  Schulman \etal \cite{schulman2016learning} proposed a method based on non-rigid trajectory transfer for adapting the demonstrated trajectory from the training geometry to the test geometry, enabling the robot to tie different types of knots in the rope automatically. Chu \etal \cite{chu2016learning} learned different haptic affordances by demonstrating learning to provide a robot with an example of successful interaction with a given object. Aleotti \etal \cite{aleotti2011part} performed the task of robot grasp planning well by combining the limited and automatic 3D shape segmentation of human demonstrators for object recognition and semantic modeling. Fang \etal \cite{demo2vec2018cvpr} proposed to learn from online product review videos and then transfer the knowledge from the videos to the target image to learn the relevant regions of human-object interaction. We also consider learning object affordance from demonstrator videos but only use the action labels of the videos as supervisory signals to learn the mapping relationships from static objects to the object features interacted in the videos.

\section{Method}
\subsection{Problem Description}
In this paper, our goal is to learn the affordance of objects through the demonstration videos while only using the action class as supervision without pixel-level labels. The complete process is shown in Fig. \ref{FIG:2}. Before learning affordance, we first detect the hand in each frame of the video. Some detection results are shown in Fig. \ref{FIG:3} and then introduce a specific selection network. It combines the clues provided by the position and action of the hand to select the keyframes where the hand interacts with the objects. After preprocessing the video datasets, we introduce a Hand-aided Affordance Grounding Network (HAG-Net), as shown in Fig. \ref{FIG:5}. In the affordance learning phase (train phase), we leverage the affordance cues of the object from the demonstration videos and transferring them to a static object. Finally, in the affordance grounding phase (test phase), we feed an object image and an action label into the Hand-aided Affordance Grounding Network (HAG-Net) and output a heatmap of the region on the object where the action may appear using visualization techniques.

\subsection{Pre-processing}
\label{3.1}

Since human-object interactions frequently occur in the demonstration videos, the hand position and actions can offer valuable clues for locating the affordance region of interest. Both OPRA \cite{demo2vec2018cvpr} and EPIC \cite{Damen2018EPICKITCHENS} have no hand annotations, however, the large amount of existing work on object detection \cite{yolov3,chen2020recursive,ren2016faster} allows us to accurately detect the human hand. We choose the YOLOv3 pruning model \cite{yolov3,Liu2017learning} trained on the Oxford hand dataset \cite{DBLP:conf/bmvc/MittalZT11} to detect the hands in each frame of the two datasets. At most, two hands can be detected in each frame. If the number of detected hands is greater than two, we select the two hands with the highest confidence. Some of the detection results are shown in Fig. \ref{FIG:3} (a). As can be seen, the YOLOv3 model can accurately detect the hand.

\begin{figure*}[t]
	\centering
		\includegraphics[width=0.99\linewidth]{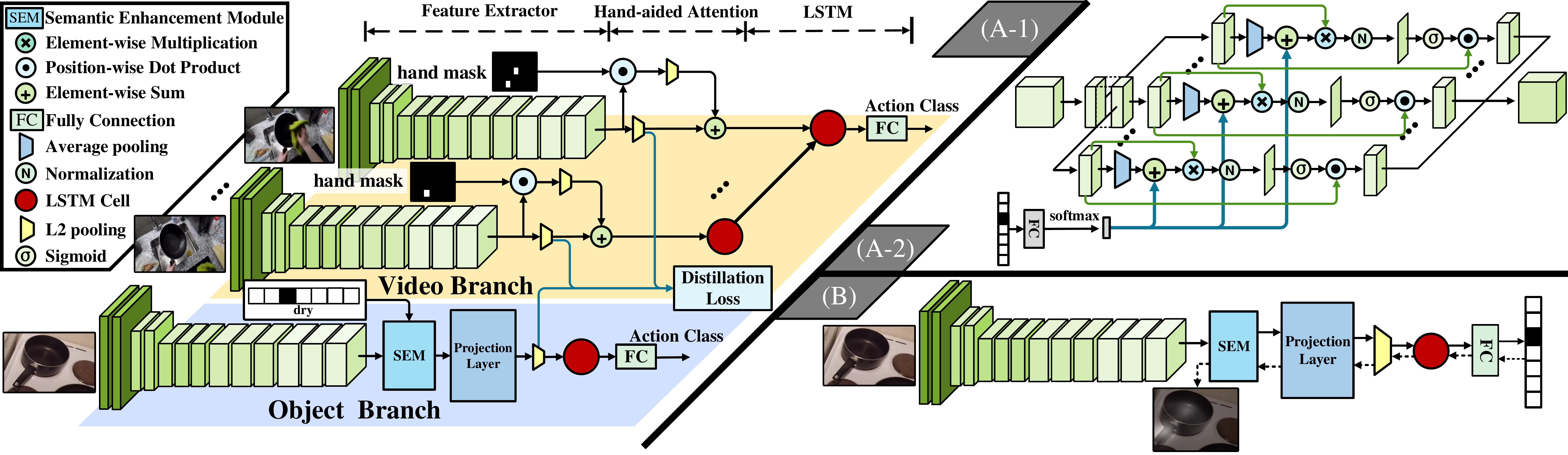}
	\caption{\textbf{The architecture of Hand-aided Affordance Grounding Network (HAG-Net).} (A) Learning object affordance from demonstration videos. (A-1) For video branches, we first extract features using ResNet50 \cite{DBLP:conf/cvpr/HeZRS16}, introduce Hand-aided attention to enhance the hand-related region features, and finally use LSTM to aggregate the features of video frames to obtain action classification results. For the object branch, HAG-Net uses the ResNet50 to extract features, which together with the action label are fed into the semantic enhancement module (SEM). Then, a projection layer transfers the affordance information from the demonstration video to the object image. Finally, we use the distillation loss to align the features of the object branch and those of the video branch. (A-2) The Semantic Enhancement Module (SEM) utilizes a combination of action categories and object apparent features to guide the network to enhance features in areas relevant to action interactions. (B) Affordance grounding. Given an object image and an action label, we use the Grad-CAM to activate the affordance regions of the feature map according to the action (affordance) label.}
	\label{FIG:5}
\end{figure*}

Since many video frames do not contain helpful information about the interaction between the human and the objects, which may cause an undesirable influence on the affordance grounding, therefore, we use a specific hand-related selection network to select the keyframes where the hand interacts with the object and filter out the frames whose intermediate states cannot provide affordance clues. The framework has a similar structure to the video branch in Fig. \ref{FIG:5} (A-1). Since hand position and action cues play an essential role in affordance grounding, we also introduce hand-aided attention in the keyframe selection process to make the network pay attention to hand-aided information. We send the feature maps after CNN into hand-aided attention to enhance the relevant features of human hands, thus making the selected frames more discriminative. We will introduce how hand-aided attention can make use of hand information better in Section \ref{hand-aided}. To capture the action information of the hands in the demonstration videos, we use the Long Short-Term Memory (LSTM) \cite{hochreiter1997long} to aggregate the temporal features. After training, we process the videos in both OPRA \cite{demo2vec2018cvpr} and EPIC \cite{Damen2018EPICKITCHENS} datasets. Specifically, we randomly choose a starting position and select eight consecutive frames, then feed into the network, and the frame with the highest confidence is selected. We only keep the frame when the action prediction is correct or the confidence is greater than a certain threshold (generally set to 0.3). We repeat this process and select several keyframes for each video. In this way, the retained frames are discriminative and can provide affordance-related information.

\subsection{Affordance Learning}
\label{3.2}
After preprocessing the video, we introduce the Hand-aided Affordance Grounding Network (HAG-Net). It is mainly divided into two parts, as shown in Fig. \ref{FIG:5}. The sub-figure (A) describes the process of affordance learning of objects from demonstration videos, and the sub-figure (B) describes the process of affordance grounding. This section mainly introduces the process of affordance learning, and we will describe the affordance grounding in Section \ref{3.4}. We divide the network into two branches during the training phase to learn how human interacts with the objects in the videos and transfer the knowledge to the object branch. 

\par For the video branch, let a video contain T frames $\mathcal{V}=\{f_{1},f_{2},...,f_{T}\}$ with an afforded action class $c$. First, we use a ResNet50 \cite{DBLP:conf/cvpr/HeZRS16} backbone to extract the features of each frame, and the features of the video are represented as $\mathcal{X}=\{X_{1},X_{2},...,X_{T}\}$. We then input $\mathcal{X}$ into hand-aided attention (in Section \ref{hand-aided}) to obtain the enhanced feature $\mathcal{M}=\{M_{1},M_{2},...,M_{T}\}$. After enhancing the relevant features of the human hand, we input them into the LSTM \cite{hochreiter1997long} for aggregating the hand action features:
\begin{equation}
    H(\mathcal{V})=LSTM(M_{1},M_{2},...,M_{T}), \label{eq:no5}
\end{equation}
where $H$ is the feature representation after aggregation. Finally, $H$ is sent to a classifier to predict the action class. 

\par For the object branch, we first extract the features of the object image using the same ResNet50 backbone. Then, the semantic enhancement module $SEM(\cdot)$ is introduced to enhance the features. Since there is a large gap between the feature of the interacting object and the static object, we use a projection layer $Proj(\cdot)$ to learn the object feature representation $\tilde{X}$ when the interaction occurs:
\begin{equation}
    \tilde{X}=Proj(SEM(X,c)), \label{eq:no6}
\end{equation}
$X$ is the feature map of the object image after ResNet50 \cite{DBLP:conf/cvpr/HeZRS16}.

Finally, we introduce a distillation loss ($\mathcal{L}_{distill}$) to align the feature space of static objects to the feature space of human-object interaction, from the affordance of the objects in the demonstration video to the static objects. To ensure that the feature representation of the object after the projection layer has the same classification performance as the video branch, we send the output of the projection layer to the same LSTM and action classifier and calculate the classification loss of the object branch. The total loss ($\mathcal{L}_{total}$) is mainly composed of the following three parts:
\begin{equation}
   \mathcal{L}_{total}=\lambda_{vcls}\mathcal{L}_{vcls}+\lambda_{distill}\mathcal{L}_{distill}+\lambda_{ocls}\mathcal{L}_{ocls},  \label{eq:no18}
\end{equation}
where $\lambda_{vcls}$, $\lambda_{distill}$, and $\lambda_{ocls}$ are hyper-parameters that balance the loss items. $\mathcal{L}_{vcls}$ and $\mathcal{L}_{ocls}$ represent the cross-entropy loss of video branch and object branch for action classification, respectively. 
In the following sections, we present the details of \textbf{Hand-Aided Attention}, \textbf{Semantic Enhancement Module}, \textbf{Projection Layer}, and \textbf{Distillation Loss}, respectively.

\myPara{Hand-Aided Attention.}
\label{hand-aided}
We use the results of Section \ref{3.1} to enhance the features of the hand regions. Since we need the context information of the hands and the objects, we expand the bounding box of the detected hand and exclude the hand regions that may occlude the object. The expansion and exclusion regions are controlled by parameters $\alpha_{1}$ ($0<\alpha_{1}<1$) and $\alpha_{2}$ ($0<\alpha_{2}<1$). As shown in Fig. \ref{FIG:3} (b), $h$ and $w$ are the height and width of the hand bounding box, respectively. The expansion region is the bounding box enclosed by $(1+\alpha_{1})\times w$ and $(1+\alpha_{1})\times h$, while the exclusion region is the bounding box enclosed by $(1-\alpha_{2})\times w$ and $(1-\alpha_{2})\times h$. The hand mask ($Mask_t$) used in our network is shown in the right part of Fig. \ref{FIG:3} (b). We set $\alpha_{2}$ to 1 when calculating the hand-aided attention in the selection network and do not exclude the hand region. While in the affordance learning phase, since it will affect the extraction of affordance clues, we set $\alpha_{2}$ to 0.4 accordingly. After obtaining $Mask_t$, we calculate the masked feature $X_{t}^M$ by multiplying it with the feature map $X_{t}$ of the video frame via an position-wise dot product operator $\odot$:
\begin{equation}
    X_{t}^M=X_{t} \odot Mask_{t}. \label{eq:no1}
\end{equation}

Then the feature representation of each frame and the mask of the hand are respectively passed through L2 pooling layer ($L2P(\cdot)$) and added to obtain the enhanced feature $M_t$:
\begin{equation}
    g_{t}=L2P(X_{t}), \label{eq:no2}
\end{equation}
\begin{equation}
    g_{t}^M=L2P(X_{t}^M), \label{eq:no3}
\end{equation}
\begin{equation}
    M_{t}=g_{t}+g_{t}^M, \label{eq:no4}
\end{equation}
where $g_{t}$ and $g_{t}^M$ are the feature representations of the video frame and the masked feature after pooling, respectively.

\label{SEM}
\myPara{Semantic Enhancement Module.}
 
Many previous studies \cite{singh2019hetconv,sun2019high,wu2018group} have demonstrated that the features learned by the grouping mechanism are more compact and have better representational power. Therefore, we introduce an Semantic Enhancement Module (SEM) to improve the feature representation capability of object branches by leveraging a group-wise feature enhancement mechanism. Furthermore, within each grouping, the feature of the object image is modulated jointly by action categories and global features, thus enabling the network to focus more on affordance-related features. Our SEM is shown in Fig. \ref{FIG:5} (A-2), the feature map of size $C \times H \times W$ is divided into G groups along the channel dimension. The feature of the $i$th position of the $k$th group is represented as $X_i^k$, $k \in [1,G], i \in [1,H\times W]$. 

Besides, we use a fully connected layer $fc(\cdot)$ to map the one-hot action label to have $C$ channels and use a Softmax layer to get the action class weights $X_{class}$:
\begin{equation}
    X_{class}=\text{Softmax}(fc(\text{one-hot}(c))), \label{eq:no7}
\end{equation}
where $\text{one-hot}(\cdot)$ is the one-hot representation of action class. We also divide $X_{C}$ into G groups.
Then, we use a global average pooling to aggregate the global statistical feature from $X^{k}$ and add it with $X_{class}^{k}$ to obtain the semantic vector $S^{k}$ of the $k$th group:
\begin{equation}
    S^{k}=\frac{1}{H \times W}\sum_{i=1}^{H \times W} X_{i}^k+X_{class}^k. \label{eq:no8}
\end{equation}

Then, we calculate the correlation coefficient $c^k$ between $X^k$ and $S^k$ and normalize it at each position as follows:
\begin{equation}
\begin{split}
    & c^k=S^k \otimes X^k, \\
\end{split}
\end{equation}
\begin{equation}
   \hat{c}_{i}^k=\frac{c_{i}^k-\mu^{k}}{\sigma^{k}+\epsilon}, \label{eq:no10}
\end{equation}
\begin{equation}
   \mu^{k}=\frac{1}{H \times W}\sum_{j=1}^{H \times W} c_{j}^k,  \label{eq:no11}
\end{equation}
\begin{equation}
   \sigma^{k} = \sqrt{\frac{1}{H \times W}\sum_{j}^{H \times W} (c_{j}^k-\mu^{k})^2}, \label{eq:no12}
\end{equation}
where $\otimes$ means the element-wise multiplication operation, $\mu^{k}$ and $\sigma_{k}$ represent the mean and standard deviation of the group of coefficients, respectively, and $\epsilon$ is a regularization constant. After that, we introduce $\gamma$ and $\beta$, which scale and shift the normalized value $\hat{c}_{k, i}$:
\begin{equation}
   a_{i}^k=\gamma^{k}\hat{c}_{i}^k+\beta^{k}, \label{eq:no13}
\end{equation}
where $k$ denotes the $k$th group of parameters. Finally, we input $a_{i}^k$ to a sigmoid function gate $\sigma(\cdot)$, and multiply it with the original feature representation $X_{i}^k$ to get the enhanced feature representation $\hat{X}_{i}^k$:
\begin{equation}
    \hat{X}_{i}^k=X_{i}^k\odot \sigma(a_{i}^k).  \label{eq:no14}
\end{equation}

\label{projection}
\myPara{Projection Layer and Distillation Loss.}

We introduce a projection layer and a distillation loss to align the feature representations of the object branch and the video branch, reducing the distance in feature space between the object in the static image and the object in the scene of human-object interaction in the video. During aligning the feature representation of interaction between humans and objects to that of static objects, we mainly focus on two factors: 1) the highest-confidence frame $X_{t^{*}}$ in the input video contains the affordance clues of the hand's position, and 2) the average feature of the input contains the hand context. 

Specifically, in the process of obtaining the features of $X_{t^{*}}$, we determine $t^{*}$ using the following equation:
\begin{equation}
     t^{*}=\mathop{\min\arg}\limits_{t \in 1,...,T}\mathcal{L}_{fcls}(LSTM(M_{1},...,M_{T}),c),  \label{eq:no15}
\end{equation}
where $L_{fcls}$ is the cross-entropy loss for classification using the hidden state of LSTM at time $t$ (each frame of input). The average feature of the video is represented as $\bar{X}_{\mathcal{V}}$:
\begin{equation}
   X_{\mathcal{V}}=\frac{1}{T}(\sum_{t=1}^T X_{t}).  \label{eq:no16}
\end{equation}

We define a distillation loss ($\mathcal{L}_{distill}(\tilde{X},X_{\mathcal{V}}, X_{t^{*}})$) to adjust the projection layer so that it can align the feature representation of static object and that of interactive object:
\begin{equation}
\begin{split}
   \mathcal{L}_{distill}&=\lambda_{1}\times||L2P(\tilde{X})-L2P(X_{t^{*}})||_{2} \\
                                      & +\lambda_{2}\times||L2P(\tilde{X})-L2P(X_{\mathcal{V}})||_{2},  \label{eq:no17}
\end{split}
\end{equation}
where $L2P(\cdot)$ is the L2 pooling operation, $\lambda_1$ and $\lambda_2$ are the loss weights. $\mathcal{L}_{distill}$ is the key to ensure that the object image is mapped to the feature space of character interactions. 

\subsection{Affordance Grounding}
\label{3.4}
\par In Affordance Grounding process, the input is just a static object image, and our goal is to infer the interaction regions of all possible actions on this object image. For each action class $c$, the model generates a heatmap of the position where the human interacts with the object, as shown in Fig. \ref{FIG:5} (B). In this paper, we use the Grad-CAM \cite{selvaraju2017grad} feature visualization technique to establish the mapping between action labels to regions related to human-object interaction. In the ablation study in Section \ref{ablation}, we also compare other advanced visualization methods and prove that Grad-CAM can achieve the best results. The detailed approach is as follows: 1) for a specific object image encoding $X$ and action class $c$, the sensitivity of the action relative to each channel of the last layer of the feature map is calculated, which can be regarded as the attention mask of the feature map of this layer; 2) unlike Grad-CAM, we directly combine it with the weighted linear combination of the last layer of the feature map instead of taking the gradient map to the mean value; and 3) we input it to the ReLU activation function for further processing. Using the ReLU function can help our model focus on the pixels that have a positive impact on a particular category:
\begin{equation}
   H_{c}(X)=\sum_{k}ReLU(\frac{\partial Y^{c}}{\partial X^{k}}\odot X^{k}),  \label{eq:no19}
\end{equation}
where $\odot$ is the element-wise multiplication operator,  $X^{k}$ is the $k$th channel of the input embedding, and $\frac{\partial Y^{c}}{\partial X^{k}}$ is a two-dimensional attention mask. $H_{c}(X)$ represents the final interaction heatmap for the given action class.

\section{Experiments}
In our experiments, we explore the following questions: 
\begin{itemize}
    \item [$\bm{-}$]
    $\bm{Q1:}$ \textbf{Does our method outperform other weakly or fully supervised methods on the affordance grounding task?} (In Section \ref{results_analysis})
    \item [$\bm{-}$]
    $\bm{Q2:}$ \textbf{Does our method have excellent generalization performance on unseen objects?} (In Section \ref{results_analysis})
    \item [$\bm{-}$]
    $\bm{Q3:}$ \textbf{What is the influence of each module of our model and different visualization strategies on affordance grounding?} (In Section \ref{ablation})
\end{itemize}
\par We chose the OPRA \cite{demo2vec2018cvpr} dataset for the third-person perspective, and the EPIC \cite{Damen2018EPICKITCHENS} dataset for the first-person perspective to evaluate the effectiveness of the model in terms of both objective metrics as well as subjective visualization. In the ablation study, we explore different visualization strategies and investigate the impact of each module of our model on affordance grounding.

\subsection{Experimental Setup}

\myPara{Datasets.}
Our main goal is to use human demonstration videos to learn how people interact with objects. We need datasets that contain a large number of demonstration videos in which people interact with various objects. To this end, we conducted our experiments on the following two datasets.

\begin{itemize}

\item [$\bm{-}$] 
\textbf{Online Product Review dataset for Affordance (OPRA)} \cite{demo2vec2018cvpr}: Fang \etal \cite{demo2vec2018cvpr} proposed the OPRA dataset, which aims to use demonstration videos for object affordance inference. Each sample contains a video, an object image, affordance class, and annotations of the interacting regions on the object image.

\item [$\bm{-}$] 
\textbf{EPIC-KITCHENS (EPIC)} \cite{Damen2018EPICKITCHENS}: The dataset contains a large number of egocentric videos of activities in kitchens. Each clip contains an action label and an object. Moreover, each frame has a bounding box of the object that interacts with the person. In this paper, we use the data annotated in \cite{interaction-hotspots}.

\end{itemize}

\myPara{Evaluation Metrics.}
To evaluate the results of different models comprehensively, we choose four metrics from \cite{bylinskii2018different}, including KLD \cite{bylinskii2018different}, SIM \cite{swain1991color}, AUC-J \cite{DBLP:conf/iccv/JuddEDT09}, NSS \cite{peters2005components}.

\begin{table*}[!t]
  \begin{center}
  \small
  \renewcommand{\arraystretch}{1.}
  \renewcommand{\tabcolsep}{2.8pt}
   \caption{\textbf{Results of the proposed method and several weakly supervised methods}, including Saliency detection methods (Egogaze \cite{DBLP:conf/eccv/HuangCLS18}, Mlnet \cite{DBLP:conf/icpr/CorniaBSC16}, DeepgazeII \cite{DBLP:journals/corr/KummererWB16}, Salgan \cite{Pan_2017_SalGAN}) and Hotspot \cite{interaction-hotspots}, and two affordance detection methods using mask labels as strong supervision during training, including Demo2vec \cite{demo2vec2018cvpr} and Affsegnet \cite{interaction-hotspots}, on the OPRA \cite{demo2vec2018cvpr} and EPIC \cite{Damen2018EPICKITCHENS} datasets. $K$, $S$, $A$, and $N$ represent the four evaluation metrics KLD \cite{bylinskii2018different}, SIM \cite{swain1991color}, AUC-J \cite{DBLP:conf/iccv/JuddEDT09}, and NSS \cite{peters2005components}, respectively. $\uparrow$ /$\downarrow$ indicates higher/lower results are better. 
   }
   \label{Table:1}
  \begin{tabular}{r|cccc|cccc|cccc|cccc}
    \hline\toprule
    \multicolumn{1}{c|}{\quad} & \multicolumn{8}{c|}{\textbf{Affordance Grounding}} & \multicolumn{8}{c}{\textbf{Generalization to Novel Objects}} \\ 
    \hline
    \multicolumn{1}{c|}{Dataset} & \multicolumn{4}{c|}{OPRA \cite{demo2vec2018cvpr}} & \multicolumn{4}{c|}{EPIC \cite{Damen2018EPICKITCHENS}} & \multicolumn{4}{c|}{OPRA \cite{demo2vec2018cvpr}} & \multicolumn{4}{c}{EPIC \cite{Damen2018EPICKITCHENS}}\\ 
    \hline
    \multicolumn{1}{c|}{Method} & \emph{$K\downarrow$} & \emph{$S\uparrow$} & \emph{$A\uparrow$} & \emph{$N\uparrow$} & \emph{$K\downarrow$} & \emph{$S\uparrow$} & \emph{$A\uparrow$} &  \emph{$N\uparrow$} & \emph{$K\downarrow$} & \emph{$S\uparrow$} & \emph{$A\uparrow$} & \emph{$N\uparrow$} & \emph{$K\downarrow$} & \emph{$S\uparrow$} & \emph{$A\uparrow$} & \emph{$N\uparrow$} \\ 
    \hline
    Center bias & $11.132$ & $0.205$ & $0.625$ & $0.323$ & $10.660$ & $0.222$ & $0.634$	& $0.333$ & $6.281$ &	$0.244$ & $0.680$ & $0.340$ & 	$5.910$ & $0.277$ & $0.699$ & $0.344$  \\
   Egogaze \cite{DBLP:conf/eccv/HuangCLS18} & 	\cellcolor{gray!20}$2.428$ & $0.245$ & $0.646$ & 	$0.247$ &	 \cellcolor{gray!20}$2.241$ & $0.273$ & $0.614$ & $0.281$ & \cellcolor{gray!20}$2.083$ & $0.278$ &	$0.694$ & $0.244$ & \cellcolor{gray!20}$1.974$ & $0.298$ & $0.673$ & $0.280$    \\
   Mlnet \cite{DBLP:conf/icpr/CorniaBSC16}	& $4.022$ &	\cellcolor{gray!20}$0.284$	& \cellcolor{gray!35}$0.763$ &	\cellcolor{gray!35}$0.607$ & $6.116$ & \cellcolor{gray!20}$0.318$ & \cellcolor{gray!20}$0.746$ &	\cellcolor{gray!20}$0.809$ & $2.458$ & \cellcolor{gray!20}$0.316$ & \cellcolor{gray!35}$0.778$ &	\cellcolor{gray!35}$0.551$ & $3.221$ & \cellcolor{gray!20}$0.361$ & \cellcolor{gray!35}$0.799$ &	\cellcolor{gray!20}$0.831$   \\
   DeepgazeII \cite{DBLP:journals/corr/KummererWB16}	& \cellcolor{gray!50}$1.897$ & \cellcolor{gray!35}$0.296$ & \cellcolor{gray!20}$0.720$ & \cellcolor{gray!20}$0.496$ & \cellcolor{gray!50}$1.352$ & \cellcolor{gray!35}$0.394$ & \cellcolor{gray!35}$0.751$ & \cellcolor{gray!35}$0.888$ & \cellcolor{gray!35}$1.757$ & \cellcolor{gray!35}$0.318$ & \cellcolor{gray!20}$0.742$ & \cellcolor{gray!20}$0.466$ & \cellcolor{gray!35}$1.297$ & \cellcolor{gray!35}$0.400$ & \cellcolor{gray!20}$0.793$ & \cellcolor{gray!35}$0.917$   \\
   Salgan \cite{Pan_2017_SalGAN}	& $2.116$\cellcolor{gray!35} & \cellcolor{gray!50}$0.309$ & \cellcolor{gray!50}$0.769$ & \cellcolor{gray!50}$0.659$ & \cellcolor{gray!35}$1.510$ & \cellcolor{gray!50}$0.395$ & \cellcolor{gray!50}$0.774$ & \cellcolor{gray!70}$0.978$ & \cellcolor{gray!50}$1.698$ & \cellcolor{gray!50}$0.337$ & \cellcolor{gray!50}$0.790$ & \cellcolor{gray!50}$0.622$ & \cellcolor{gray!50}$1.296$ & \cellcolor{gray!70}$0.406$ & \cellcolor{gray!50}$0.808$ & \cellcolor{gray!50}$0.987$    \\
   Hotspot \cite{interaction-hotspots} & \cellcolor{gray!70}$1.427$ & \cellcolor{gray!70}$0.362$ & \cellcolor{gray!70}$0.806$ & \cellcolor{gray!70}$0.907$ & \cellcolor{gray!70}$1.258$ & \cellcolor{gray!70}$0.404$ & \cellcolor{gray!70}$0.785$ & \cellcolor{gray!50}$0.923$ & \cellcolor{gray!70}$1.381$ & \cellcolor{gray!95}$0.374$ & \cellcolor{gray!95}$0.826$ & \cellcolor{gray!70}$0.912$ & \cellcolor{gray!70}$1.249$ & \cellcolor{gray!50}$0.405$ & \cellcolor{gray!70}$0.817$ & \cellcolor{gray!70}$1.001$   \\
   \hline
   Ours & \cellcolor{gray!95}$1.409$ & \cellcolor{gray!95}$0.365$ & \cellcolor{gray!95}$0.812$ & \cellcolor{gray!95}$0.948$ & \cellcolor{gray!95}$1.209$ & \cellcolor{gray!95}$0.414$ & \cellcolor{gray!95}$0.801$ & \cellcolor{gray!95}$1.045$ & \cellcolor{gray!95}$1.366$ & \cellcolor{gray!70}$0.373$ & \cellcolor{gray!70}$0.817$	 & \cellcolor{gray!95}$0.927$ & \cellcolor{gray!95}$1.197$ & \cellcolor{gray!95}$0.412$ & \cellcolor{gray!95}$0.820$ & \cellcolor{gray!95}$1.084$    \\
   \hline
   Img2heatmap \cite{interaction-hotspots} & $1.473$ & $0.355$ & $0.821$ & $0.894$ & $1.400$ & $0.359$ & $0.794$ & $0.925$ & $1.431$ & $0.362$ & $0.820$ & $0.850$ & $1.466$ & $0.353$ & $0.770$ & $0.720$   \\

   Demo2vec\cite{demo2vec2018cvpr} & $1.197$ & $0.482$ & $0.847$ & $1.170$ & - & - & - & - & - & - & - & - & - & - & - & - \\
    \hline\bottomrule
    \end{tabular}
    \end{center}
  \end{table*}

\begin{figure*}[t]
	\centering
		\begin{overpic}[width=0.99\linewidth]{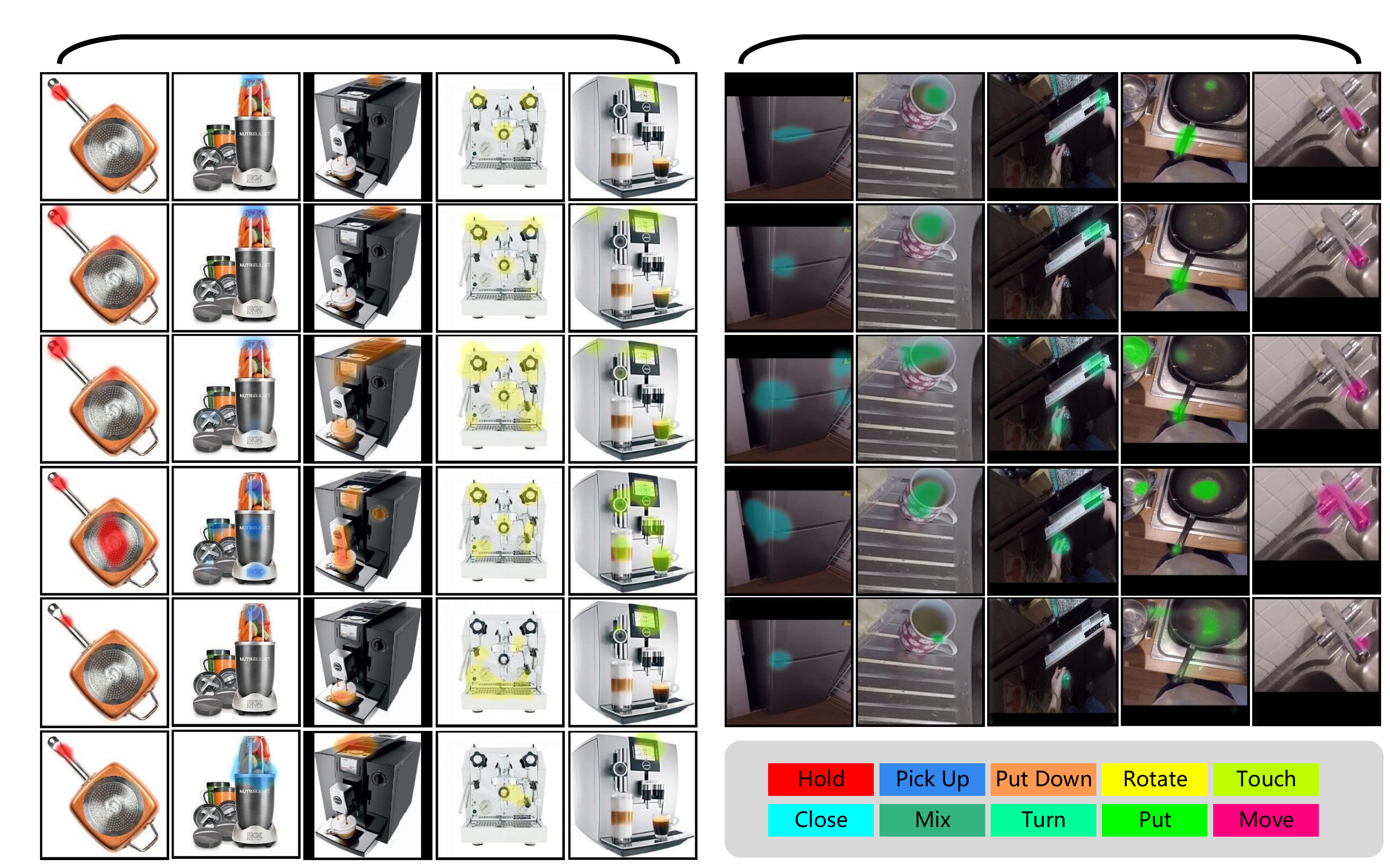}
          \put(21.5,59.3){\colorbox{white}{\textbf{OPRA} \cite{demo2vec2018cvpr}}}
          \put(71,59.3){\colorbox{white}{\textbf{EPIC} \cite{Damen2018EPICKITCHENS}}}
          \put(-0.5,52){\footnotesize \textbf{\rotatebox{90}{GT}}}
          \put(-0.5,41){\footnotesize \textbf{\rotatebox{90}{ Ours}}}
          \put(-0.5,31){\footnotesize \textbf{\rotatebox{90}{Hotspot}}}
          \put(1,31.5){\footnotesize \textbf{\rotatebox{90}{ \cite{interaction-hotspots}}}}
          \put(-0.5,22){\footnotesize \textbf{\rotatebox{90}{Saliency}}}
          \put(1,21){\footnotesize \textbf{\rotatebox{90}{ \cite{Pan_2017_SalGAN,DBLP:journals/corr/KummererWB16,DBLP:conf/icpr/CorniaBSC16,DBLP:conf/eccv/HuangCLS18}}}}
          \put(-0.5,10.5){\footnotesize \textbf{\rotatebox{90}{Img2heatmap}}}
          \put(1,13.5){\footnotesize \textbf{\rotatebox{90}{ \cite{interaction-hotspots}}}}
          \put(-0.5,1){\footnotesize \textbf{\rotatebox{90}{Demo2vec}}}
          \put(1,3){\footnotesize \textbf{\rotatebox{90}{ \cite{demo2vec2018cvpr}}}}
		\end{overpic}
		\caption{\textbf{Visualization of the affordance heatmaps generated by different methods,} including Hotspot \cite{interaction-hotspots}, Saliency (Egogaze \cite{DBLP:conf/eccv/HuangCLS18}, Mlnet \cite{DBLP:conf/icpr/CorniaBSC16}, DeepgazeII \cite{DBLP:journals/corr/KummererWB16} and Salgan \cite{Pan_2017_SalGAN}), Img2heatmap \cite{interaction-hotspots} and Demo2vec \cite{demo2vec2018cvpr}.
		}
	\label{FIG:6}
\end{figure*}

\begin{itemize}

\item [$\bm{-}$]
\textbf{KLD }\cite{bylinskii2018different}: Kullback-Leibler Divergence (KLD) is used to measure the distribution difference between the prediction map and the target map. Given a prediction map $P$ and a ground truth map $Q^{D}$, $KL(\cdot)$ is computed as follows:
\begin{equation}
   KL(P,Q^{D})=\sum_{i}Q_{i}^{D}log(\epsilon + \frac{Q_{i}^{D}}{\epsilon+P_{i}}), \label{eq:no20}
\end{equation}
where $\epsilon$ is a regularization constant.

\item [$\bm{-}$]
\textbf{SIM }\cite{swain1991color}: The similarity metric (SIM) measures the similarity between the prediction map and the ground truth map. Given a prediction map $P$ and a continuous ground truth map $Q^{D}$, $SIM(\cdot)$ is computed as the sum of the minimum values at each pixel, after normalizing the input maps:
\begin{equation}
\begin{split}
   &SIM (P,Q^{D})=\sum_{i}min(P_{i},Q_{i}^{D}),\\
    & where\quad \sum_{i}P_{i}=\sum_{i}Q_{i}^{D}=1. \label{eq:no21}
\end{split}
\end{equation}

\item [$\bm{-}$]
\textbf{AUC-J }\cite{DBLP:conf/iccv/JuddEDT09}: AUC-Judd (AUC-J) is a variant of AUC proposed by Judd \etal \cite{DBLP:conf/iccv/JuddEDT09}. It measures the relative prediction map values at ground truth locations.

\item [$\bm{-}$]
\textbf{NSS }\cite{peters2005components}: The Normalized Scanpath Saliency measures the correspondence between the prediction map and the ground truth, and it treats false positives and false negatives symmetrically. Given a prediction map $P$ and a binary ground truth map $Q^{D}$, $NSS(\cdot)$ computes the average normalized prediction at ground truth locations:
\begin{equation}
\begin{split}
   &NSS(P,Q^{D})=\frac{1}{N}\sum_{i}\hat{P}\times Q_{i}^{D},\\
    & where\quad N=\sum_{i}Q_{i}^{D}\quad and \quad \hat{P}=\frac{P-\mu(P)}{\sigma(P)}. \label{eq:no22}
\end{split}
\end{equation}

\end{itemize}

\myPara{Implementation Details.}
Each video in the OPRA \cite{demo2vec2018cvpr} dataset has an object image associated with it, while the EPIC \cite{Damen2018EPICKITCHENS} dataset does not. Thus, for the EPIC dataset, we use the provided bounding box to crop the object from the video frame according to \cite{interaction-hotspots} and randomly select an image whose class matches the object class in the video. Due to the background of EPIC is more complex and the apparent features between the objects in the video and the objects in the object image are different, we replace L2 loss with triplet loss when calculating $\mathcal{L}_{distill}(\tilde{X},\tilde{X^{'}},X_{\mathcal{V}}, X_{t^{*}})$. The triplet can better utilize positive and negative samples for feature learning so that the same category is close in the space and different categories are far away in the space. Closing the gap of features belonging to the same category can help the network better focus on interacting objects and ignore irrelevant backgrounds. The $\mathcal{L}_{distill}$ is calculated as follows:
\begin{equation}
\begin{split}
   \mathcal{L}_{distill}=&\lambda_{1}\times \max [0,d(L2P(X_{t^{*}}),L2P(\tilde{X}))\\
                                                &-d(L2P(X_{t^{*}}),L2P(\tilde{X^{'}}))+M] \\
                                                &+\lambda_{2}\times \max [0,d(L2P(X_{\mathcal{V}}),L2P(\tilde{X})) \\
                                                &-d(L2P(X_{\mathcal{V}}),L2P(\tilde{X^{'}}))+M]  \label{eq:no23}.
\end{split}
\end{equation}
\par We set $\lambda_{vcls}=\lambda_{ocls}=1$ on the two datasets. For OPRA dataset, we set $\lambda_{1}=1$, $\lambda_{2}=0.2$, and $\lambda_{distill}=0.1$. For EPIC dataset, we set $\lambda_{1}=1$, $\lambda_{2}=0.5$, and $\lambda_{distill}=1$. We carry out all experiments on 1080Ti, and set the batch size and learning rate to 128 and 1e-4, respectively. During training the hand-related selection network, the video input is eight frames, while during affordance learning from demonstration videos, the video input is three frames. Furthermore, we set the stride of the last two residual stages of ResNet50 to 1 and using dilated convolutions in the convolutional layers. In this way, the output spatial resolution of ResNet50 is 1/8 of the input.

\subsection{Contenders}
We compare the performance of our method with several state-of-the-art methods on OPRA \cite{demo2vec2018cvpr} and EPIC \cite{Damen2018EPICKITCHENS}. It is worth to notice that we select a series of saliency detection models as the comparison methods, because the human visual system has the ability to quickly orient attention to the most informative parts of visual scenes and the research on salient object detection is derived from this ability of the human visual system to extract the most attention-grabbing objects from the image. These saliency detection models include Egogaze \cite{DBLP:conf/eccv/HuangCLS18}, Mlnet \cite{DBLP:conf/icpr/CorniaBSC16}, DeepgazeII\cite{DBLP:journals/corr/KummererWB16}, and Salgan\cite{Pan_2017_SalGAN}. The generated heatmaps by them represent the parts of the object first noticed by the human visual system. We directly use the trained saliency detection models for test. In addition, we also choose the Hotspot \cite{interaction-hotspots}, Demo2vec \cite{demo2vec2018cvpr}, and Img2heatmap \cite{interaction-hotspots} models, where our model, Hotspot, and the saliency detection models are all weakly supervised, while Demo2vec and Img2heatmap are fully supervised. All these methods are briefly described as follows: 

\begin{itemize}

\item [$\bm{-}$] 
\textbf{Center bias}: It generates a gaussian heatmap at the center of an image. Thus, it is a simple baseline for a dataset with features of central bias. 

\item [$\bm{-}$] 
\textbf{Egogaze}\cite{DBLP:conf/eccv/HuangCLS18}: It is a hybrid gaze prediction model that exploits both the visual saliency of bottom-up and task-dependent attention transition and is the first work to explore the attention transition model in the egocentric gaze prediction task and achieves state-of-the-art results in gaze prediction.

\item [$\bm{-}$] 
\textbf{Mlnet}\cite{DBLP:conf/icpr/CorniaBSC16}: Unlike previous works that predict saliency maps directly from the last layer of convolution neural network, the model fuses features extracted from different layers of the CNN. Their method contains three main blocks: feature extraction CNN, feature encoding network (weighting of low and high features), and a prior learning network. The method achieves promising results in all datasets for saliency detection.

\item [$\bm{-}$]
\textbf{DeepgazeII}\cite{DBLP:journals/corr/KummererWB16}: Unlike other saliency models, DeepGazeII does not perform additional fine-tuning of the VGG features and only trains some output layers to predict saliency on top of VGG.

\item [$\bm{-}$]
\textbf{Salgan}\cite{Pan_2017_SalGAN}: It introduces the adversarial training mechanism of GAN for salient object prediction, which consists of two main parts: one predicts saliency maps based on the input image, and the other discriminates whether the input is the prediction result or the grounding truth. It explores the application of GAN to salient object detection and achieves excellent results on relevant datasets. 

\item [$\bm{-}$]
\textbf{Hotspot} \cite{interaction-hotspots}: It is a weakly supervised way to learn the affordance of an object through video, and affordance grounding is achieved only through action labels. 

\item [$\bm{-}$]
\textbf{Demo2vec} \cite{demo2vec2018cvpr}: It is a fully supervised way to learn the affordance of objects by learning the coded representation in the demonstration video to predict the regions in the object image that also interact with each other. The network's input is a demonstration video and an object image, and the output is the action category of the demonstration video and the interacting regions.

\item [$\bm{-}$]
\textbf{Img2heatmap} \cite{interaction-hotspots}: The encoder is a VGG16 \cite{DBLP:journals/corr/SimonyanZ14a} pre-trained on Imagenet \cite{DBLP:conf/cvpr/DengDSLL009}. The decoder is a mirrored VGG16, where the max-pooling is replaced by the upsampling operation. The network's final output is a heatmap of the same size as the input, and the number of channels is the same as the action class.

\end{itemize}

\subsection{Results Analysis}
\label{results_analysis}
In this section, we compare the ability of different models for affordance grounding on two datasets and their generalization abilities on unseen objects, as well as discuss the compatibility of our model for datasets with first- and third-person view videos. Finally, we discuss the effectiveness of our model for handling affordance with multiple possibilities.

\myPara{Affordance Grounding Results.} 
We compare our method with the state-of-the-art methods on OPRA \cite{demo2vec2018cvpr} and EPIC \cite{Damen2018EPICKITCHENS} datasets, the results are summarized in the left part of Table \ref{Table:1}. Our method surpasses all other weakly supervised methods in all metrics and is close to the supervised Demo2vec \cite{demo2vec2018cvpr} and Img2heatmap \cite{interaction-hotspots}. It proves that our method utilizes the affordance cues provided by hand position and action can achieve promising results. We also visualize the heatmaps generated by different methods in Fig. \ref{FIG:6}. Our method generates heatmaps that are closer to ground truth than those of Hotspot and saliency detection models. And there is no large response on object parts that are not related to actions. The results on some objects are even better than those of the supervised Img2heatmap \cite{interaction-hotspots} and Demo2vec \cite{demo2vec2018cvpr} methods. It shows that our method transfers the affordance cues from the hand to the static object, which can make the network pay more attention to the regions related to the affordance while suppressing the regions unrelated to the action.

\myPara{Generalization to Novel Objects.}
To verify the generalization ability of our method on new objects, we re-divide the datasets according to \cite{interaction-hotspots}. The results are shown in Table \ref{Table:1} (right part). On EPIC \cite{Damen2018EPICKITCHENS} dataset, our model outperforms all other methods in all metrics. On OPRA \cite{demo2vec2018cvpr} dataset, our method is superior to other methods in most metrics. It demonstrates that our method can predict the region of interaction on unseen objects. This is because the cues provided by our method from hand motion and position information enable the network to pay more attention to the local details of the particular affordance class of objects, thus having better generalization performance.

\myPara{Compare the Results of Merging Two Datasets.}
To verify whether our method can perform better on a more complex multi-view dataset, we merge the OPRA \cite{demo2vec2018cvpr} and EPIC \cite{Damen2018EPICKITCHENS} datasets. The experimental results are shown in Table \ref{Table:2}. Our method is superior to other methods in all metrics, suggesting that our method can also transfer hand position and action clues to static objects in a complex dataset. This is due to the fact that our method focuses more on hand-object interactions and is not too sensitive to changes in viewpoint and can be more robust to complex backgrounds.

\par
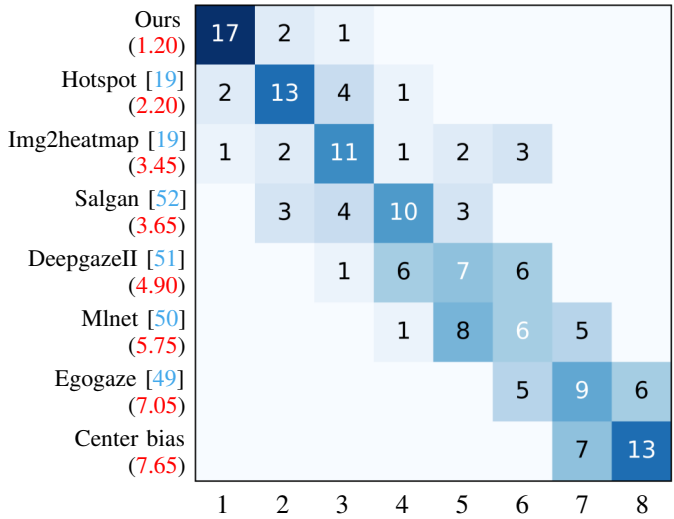
\begin{figure}[t]
  \begin{minipage}[b]{1.0\linewidth}
    \begin{center}
      \begin{overpic}[width=1\linewidth]{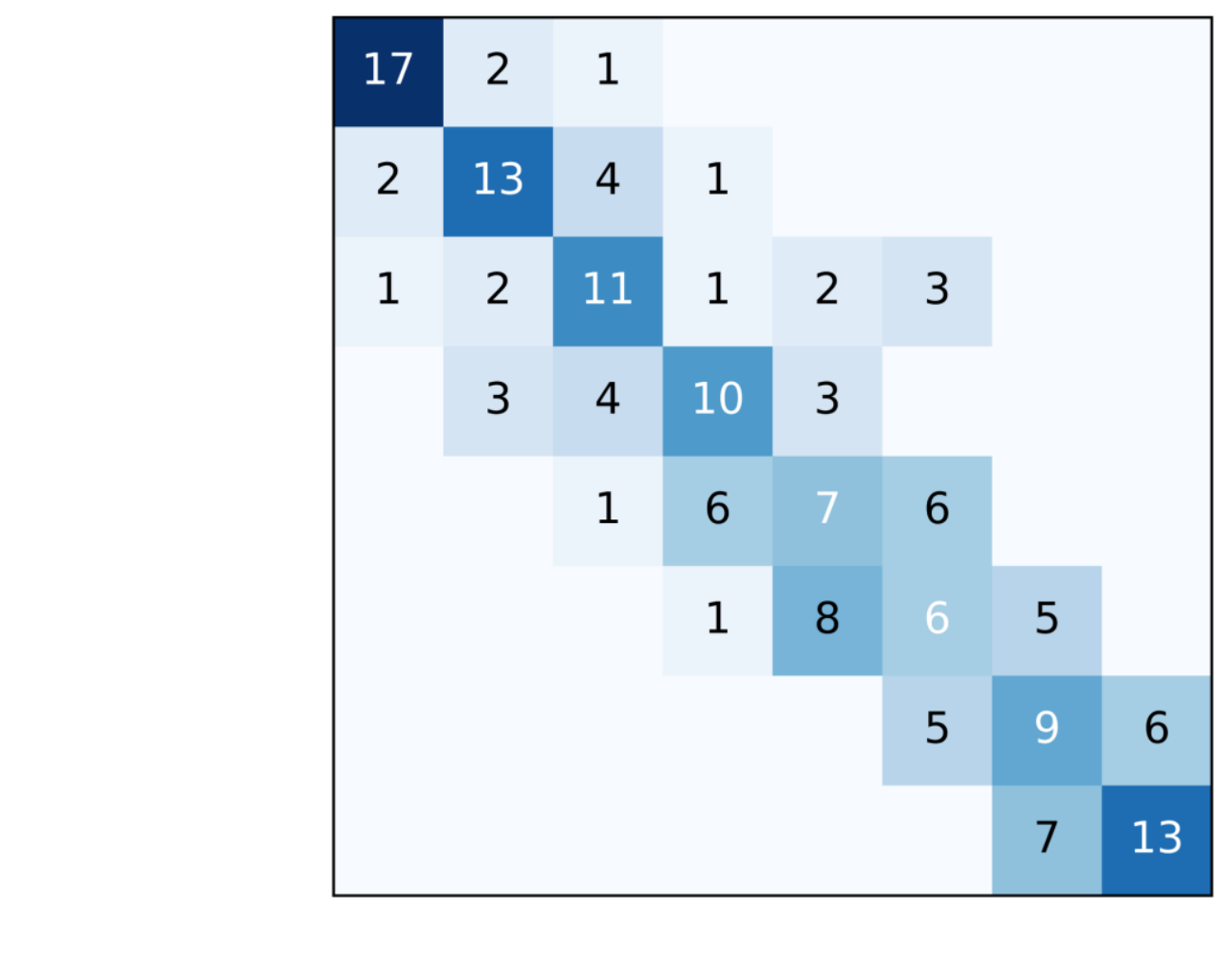}
          \put(29,2){\colorbox{white}{1}}
          \put(38,2){\colorbox{white}{2}}
          \put(47,2){\colorbox{white}{3}}
          \put(56,2){\colorbox{white}{4}}
          \put(65,2){\colorbox{white}{5}}
          \put(74,2){\colorbox{white}{6}}
          \put(83,2){\colorbox{white}{7}}
          \put(92,2){\colorbox{white}{8}}
          
          \put(18,75){\small Ours}
           \put(17,71){\small (\textcolor{red}{1.20})}
          \put(7,66){\small Hotspot \cite{interaction-hotspots} }
          \put(17,62){\small (\textcolor{red}{2.20})}
          \put(-1,57){\small Img2heatmap \cite{interaction-hotspots}}
          \put(17,53){\small (\textcolor{red}{3.45})}
          \put(9,48){\small Salgan \cite{Pan_2017_SalGAN}}
          \put(17,44){\small (\textcolor{red}{3.65})}
          
          \put(2,39){\small DeepgazeII \cite{DBLP:journals/corr/KummererWB16}}
          \put(17,35){\small (\textcolor{red}{4.90})}
          \put(10,30){\small Mlnet \cite{DBLP:conf/icpr/CorniaBSC16}}
          \put(17,26){\small (\textcolor{red}{5.75})}
          \put(6,21){\small Egogaze\cite{DBLP:conf/eccv/HuangCLS18}}
          \put(17,17){\small (\textcolor{red}{7.05})}
          \put(9,12){\small Center bias}
          \put(17,8){\small (\textcolor{red}{7.65})}
        \end{overpic}
    \end{center}
    \caption{\textbf{Rank list.} We rank the 20 results of different methods on the tasks of \textbf{``Affordance grounding''}, \textbf{``Generalization to novel object''}, and \textbf{``Merging two datasets''} in Table \ref{Table:1} and Table \ref{Table:2}, where the element ($i$, $j$) indicates how many metrics that the model $i$ are ranked the $j$th. The \textcolor{red}{red} letter denotes the average rank. 
    }
    \label{FIG:7}
  \end{minipage}
\end{figure}

 We compared the results of ``Affordance grounding'', ``Generalization to novel objects'', and `` Merging two datasets'' for three task settings with 20 metrics, ranked the results of various methods on each metric in each task setting, and summarized the result as a matrix, where each element (i,j) indicates how many metrics the model has ranked the $j$th. The results are shown in Fig. \ref{FIG:7}, from which we can see that our method achieves promising results in different task settings, with stronger robustness and generalization ability.

\begin{table}[t]
    \caption{\textbf{The results of different methods on the mixture of OPRA \cite{demo2vec2018cvpr} and EPIC \cite{Damen2018EPICKITCHENS} datasets}, $i.e.$, a dataest with images at different views.
    }
    \label{Table:2}
  \begin{center}
  \small
  \renewcommand{\arraystretch}{1.}
  \renewcommand{\tabcolsep}{8pt}
  \begin{tabular}{r|cccc}
    \hline\toprule
    \multicolumn{1}{c|}{\textbf{Dataset}} & \multicolumn{4}{c}{\textbf{OPRA} \cite{demo2vec2018cvpr} \textbf{+} \textbf{EPIC} \cite{Damen2018EPICKITCHENS}} \\ 
    \hline
    \multicolumn{1}{c|}{Method} & \emph{$K\downarrow$} & \emph{$S\uparrow$} & \emph{$A\uparrow$} & \emph{$N\uparrow$} \\
    \hline
    Center bias & $10.968$ & $0.211$ & $0.629$ & $0.327$ \\	
    Egogaze\cite{DBLP:conf/eccv/HuangCLS18} &	$2.361$ & $0.255$ & $0.635$ & $0.327$   \\
    Mlnet\cite{DBLP:conf/icpr/CorniaBSC16} & $4.735$ & $0.297$ & $0.757$ & $0.676$  \\
    DeepgazeII\cite{Pan_2017_SalGAN} & $1.705$ & $0.331$ & $0.730$ & $0.629$  \\
    Salgan\cite{DBLP:journals/corr/KummererWB16} & $1.902$ & $0.339$ & $0.771$ & $0.768$   \\
    Hotspot\cite{interaction-hotspots} & $1.435$ & $0.366$ & $0.769$ & $0.825$ \\
     \hline
     \rowcolor{mygray}
    Ours & $\normalsize\textbf{1.378}$ & $\normalsize\textbf{0.373}$ & $\normalsize\textbf{0.786}$ & $\normalsize\textbf{0.901}$   \\    
     \hline
    Img2heatmap \cite{interaction-hotspots} & 1.444 & 0.358 & 0.769 & 0.811 \\
    \hline\bottomrule
    \end{tabular}
    \end{center}
  
  \end{table}

\begin{table}[t]
\caption{\textbf{Results of different methods on twelve most frequent classes in the EPIC \cite{Damen2018EPICKITCHENS} dataset.} The best performing method is highlighted in \textbf{bold}.
}
\label{Table:3}
  \begin{center}
  \small
  \renewcommand{\arraystretch}{1.}
  \renewcommand{\tabcolsep}{8pt}
  \begin{tabular}{r|cccc}
    \hline\toprule
    \multicolumn{1}{c|}{\textbf{Dataset}} & \multicolumn{4}{c}{\textbf{EPIC} \cite{Damen2018EPICKITCHENS}} \\ 
    \hline
    \multicolumn{1}{c|}{Method} & \emph{$K\downarrow$} & \emph{$S\uparrow$} & \emph{$A\uparrow$} & \emph{$N\uparrow$} \\ 
    \hline
    Center bias &	$10.478$ & $0.219$ & $0.632$ & $0.328$ \\
    Egogaze\cite{DBLP:conf/eccv/HuangCLS18} &	$2.335$ & $0.258$ & $0.614$ & $0.328$   \\
    Mlnet\cite{DBLP:conf/icpr/CorniaBSC16} & $5.888$ & $0.315$ & $0.744$ & $0.796$  \\
    DeepgazeII\cite{Pan_2017_SalGAN} & $1.434$ & $0.373$ & $0.746$ & $0.862$  \\
    Salgan\cite{DBLP:journals/corr/KummererWB16} & $1.537$ & $0.381$ & $0.772$ & $0.968$   \\
    Hotspot \cite{interaction-hotspots} &	$1.347$ & $0.376$ & $0.781$ &	$0.905$  \\
    \hline
    \rowcolor{mygray}
    Ours   & $\normalsize\textbf{1.292}$ & $\normalsize\textbf{0.387}$ & $\normalsize\textbf{0.798}$ & $\normalsize\textbf{1.030}$ \\
    \hline
    Img2heatmap \cite{interaction-hotspots} & $1.426$ & $0.350$ & $0.791$ & $0.911$ \\
    \hline\bottomrule
    \end{tabular}
    \end{center}
\end{table}

\begin{figure}[t]
	\centering
		\begin{overpic}[width=1\linewidth]{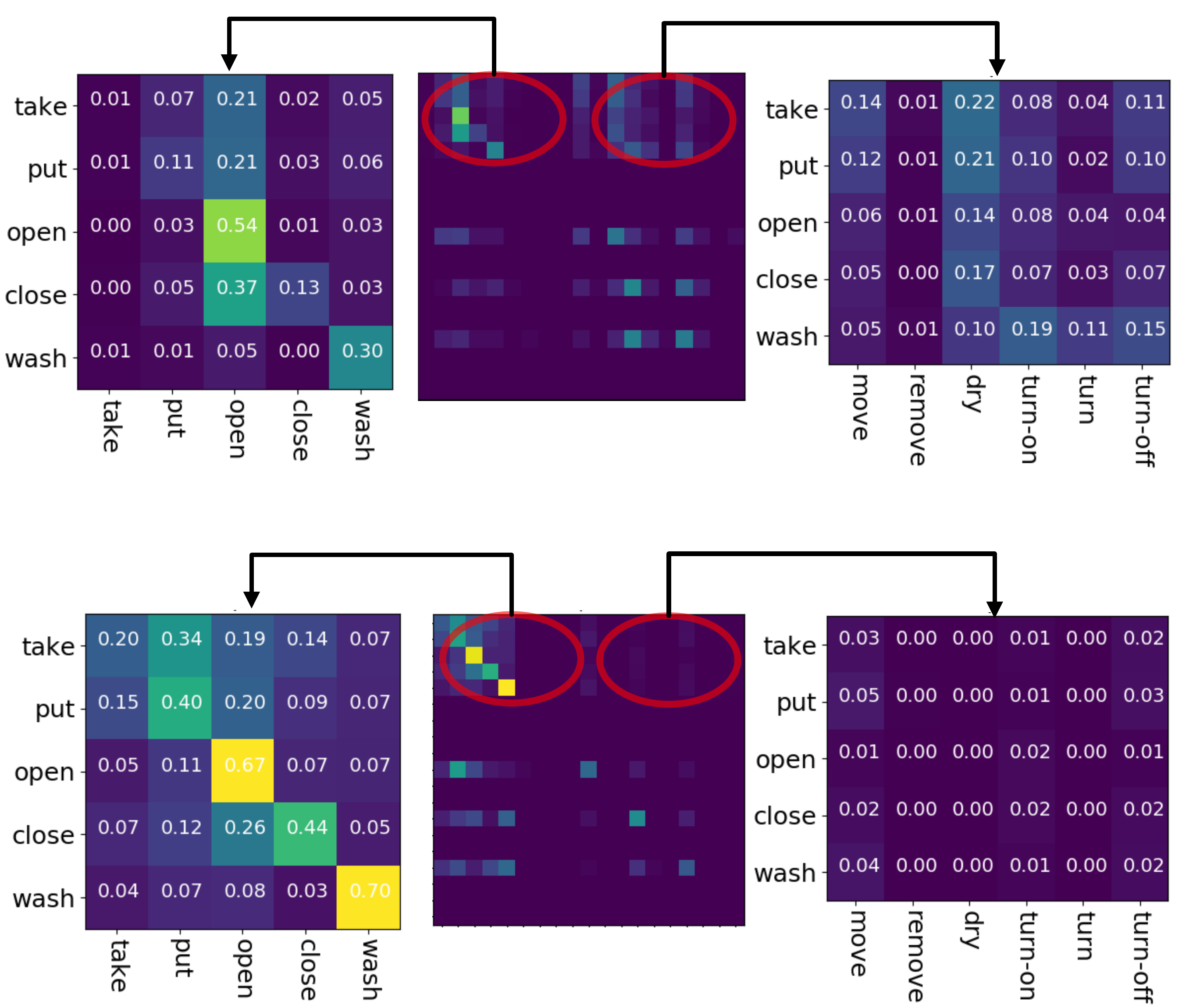}
		    \put(36, 46){\colorbox{white}{\textbf{(a) Hotspot} \cite{interaction-hotspots}}}
		    \put(42, 1){\colorbox{white}{\textbf{(b) Ours}}}
		\end{overpic}
	\caption{\textbf{The confusion matrices of Hotspot \cite{interaction-hotspots} and our method for action prediction on the EPIC dataset \cite{Damen2018EPICKITCHENS}.} We select eight most frequent classe.}
	\label{FIG:8}
\end{figure}

 \begin{figure*}[t]
	\centering
		\begin{overpic}[width=1\linewidth]{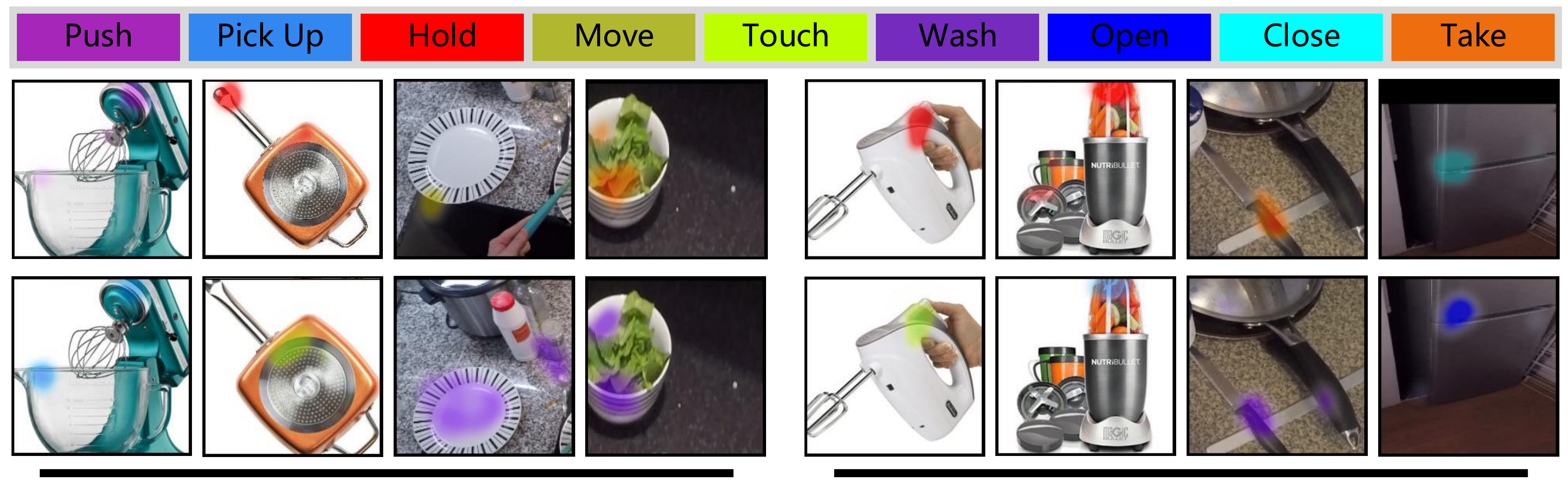}
		\put(23.5,-0.5){ \textbf{(a)}}
		\put(74,-0.5){ \textbf{(b)}}
		\end{overpic}
		\caption{\textbf{Visualization of heatmaps generated by our method when handling affordance with multiple possibilities.}
		(a) Different interactions may occur in different regions of the object. (b) Multiple possible interactions take place at the same location of the object.
		}
	\label{FIG:x}
\end{figure*}
  
\myPara{Comparison on 12 Most Frequent Classes of EPIC.}
Since the data distribution in the EPIC \cite{Damen2018EPICKITCHENS} dataset is highly unbalanced, e.g., the number of samples for some categories is less than one percent. The experimental results are more affected by the classes that contain a large number of samples. To verify the effectiveness of our method in the case of a large number of samples, we compare our method with others on the twelve most frequent classes. The experimental results are shown in Table \ref{Table:3}. Our method outperforms all other methods, implying that our method can better locate the region where people interact with objects in more samples.

\myPara{Comparison of the Action Classification Accuracy.}
Since our method and hotspot \cite{interaction-hotspots} only use action class as supervision during affordance learning, the accuracy of action recognition affects the results of affordance grounding. We compare the performance of action classification between ours and hotspot \cite{interaction-hotspots} on the EPIC \cite{Damen2018EPICKITCHENS} dataset. We only choose the eight most frequent classes and calculate the confusion matrix, as shown in Fig. \ref{FIG:8}. Our method performs better on ``open'', ``close'', ``take'', ``put'', and ``wash'', demonstrating that it can better distinguish different actions by combining hand position and action-related affordance cues to improve the result of action classification.

\begin{table*}[t]
\caption{\textbf{Comparison of different visualization approaches,} including Grad-CAM++ \cite{chattopadhay2018grad}, XGrad-CAM \cite{fu2020axiom}, and Grad-CAM \cite{selvaraju2017grad} for obtaining the object affordance region.
}
  \begin{center}
  \small
  \renewcommand{\arraystretch}{1.}
  \renewcommand{\tabcolsep}{1.pt}
  \begin{tabular}{r|cccc|cccc|cccc|cccc}
    \hline\toprule
    \multicolumn{1}{c|}{\quad} & \multicolumn{8}{c|}{\textbf{Affordance Grounding}} & \multicolumn{8}{c}{\textbf{Generalization to Novel Objects}} \\ 
    \hline
    \multicolumn{1}{c|}{Dataset} & \multicolumn{4}{c|}{OPRA \cite{demo2vec2018cvpr}} & \multicolumn{4}{c|}{EPIC \cite{Damen2018EPICKITCHENS}} & \multicolumn{4}{c|}{OPRA \cite{demo2vec2018cvpr}} & \multicolumn{4}{c}{EPIC \cite{Damen2018EPICKITCHENS}}\\ 
    \hline
    \multicolumn{1}{c|}{Method} & \emph{$K\downarrow$} & \emph{$S\uparrow$} & \emph{$A\uparrow$} & \emph{$N\uparrow$} & \emph{$K\downarrow$} & \emph{$S\uparrow$} & \emph{$A\uparrow$} &  \emph{$N\uparrow$} & \emph{$K\downarrow$} & \emph{$S\uparrow$} & \emph{$A\uparrow$} & \emph{$N\uparrow$} & \emph{$K\downarrow$} & \emph{$S\uparrow$} & \emph{$A\uparrow$} & \emph{$N\uparrow$} \\ 
    \hline
   HAG-Net \& Grad-CAM++ \cite{chattopadhay2018grad} & $1.482$ & $0.356$ & $0.770$ & $0.797$  & $1.319$ & $0.400$ & $0.750$ & $0.799$ & $1.436$ & $0.366$ & $0.773$ & $0.797$ & $1.322$ & $0.398$ & $0.758$ & $0.786$ \\
   HAG-Net \& XGrad-CAM \cite{fu2020axiom} & $1.464$ & $0.359$ & $0.782$ & $0.837$  & $1.274$ & $0.407$ & $0.775$ & $0.912$ & $1.417$ & $0.368$  & $0.785$ & $0.832$ & $1.274$ & $0.404$ & $0.785$ & $0.907$ \\
   \hline
   \rowcolor{mygray}
   HAG-Net \& Grad-CAM (Ours) & $\normalsize\textbf{1.409}$ & $\normalsize\textbf{0.365}$ & $\normalsize\textbf{0.812}$ & $\normalsize\textbf{0.948}$ & $\normalsize\textbf{1.209}$ & $\normalsize\textbf{0.414}$ & $\normalsize\textbf{0.801}$ & $\normalsize\textbf{1.045}$ & $\normalsize\textbf{1.366}$ & $\normalsize\textbf{0.373}$ & $\normalsize\textbf{0.817}$ & $\normalsize\textbf{0.927}$ & $\normalsize\textbf{1.197}$ & $\normalsize\textbf{0.412}$ & $\normalsize\textbf{0.820}$ & $\normalsize\textbf{1.084}$    \\
    \hline\bottomrule
    \end{tabular}
    \end{center}
  \label{Table:4}
  \end{table*}
  
\begin{figure*}[t]
	\centering
		\begin{overpic}[width=1\linewidth]{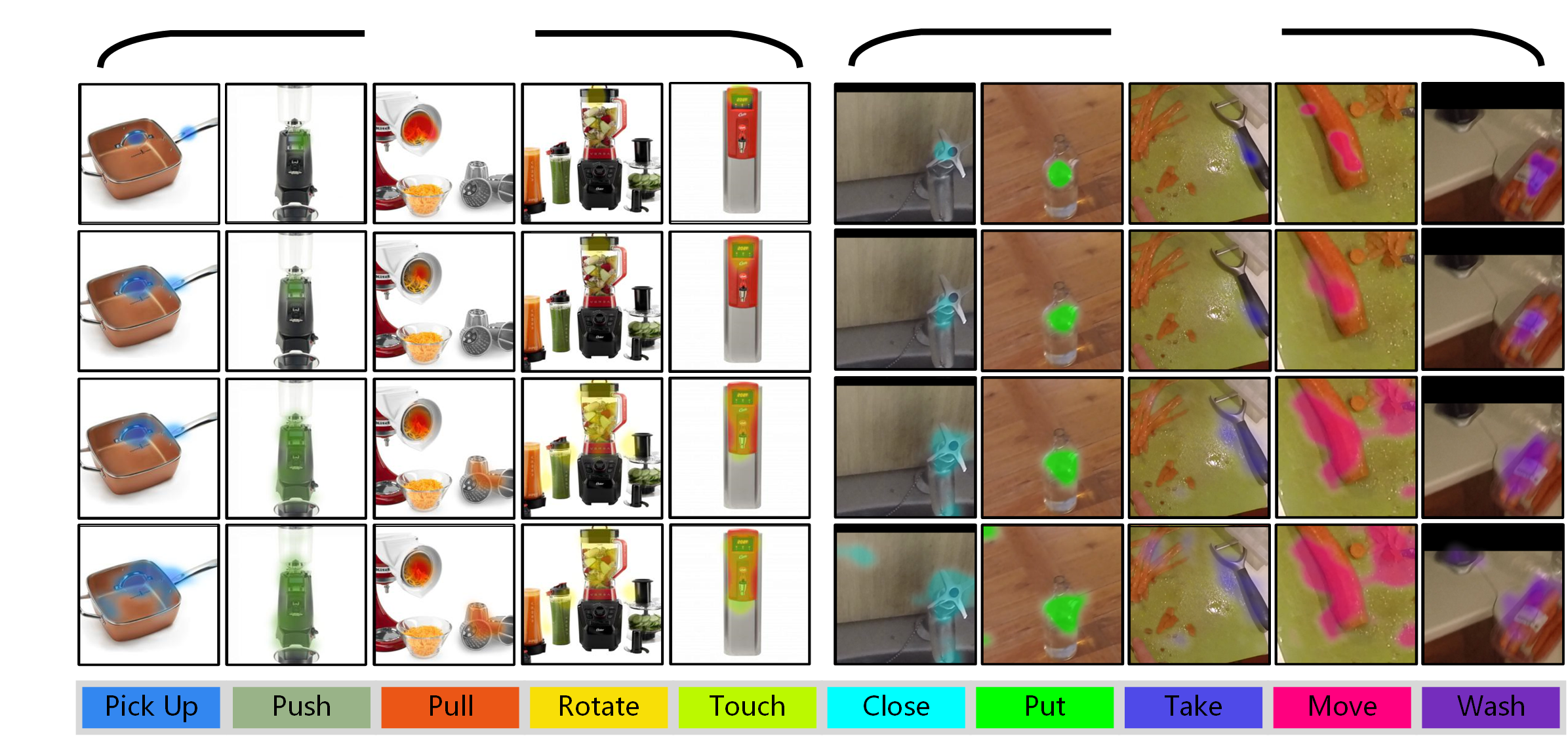}
          \put(23.6,44.4){\colorbox{white}{\textbf{OPRA} \cite{demo2vec2018cvpr}}}
          \put(71.3,44.4){\colorbox{white}{\textbf{EPIC} \cite{Damen2018EPICKITCHENS}}}
          \put(2,36){\footnotesize \textbf{\rotatebox{90}{GT}}}
          \put(0.5,24){\footnotesize \textbf{\rotatebox{90}{ HAG-Net \&}}}
          \put(2,24){\footnotesize \textbf{\rotatebox{90}{ Grad-CAM}}}
          \put(3.5,25){\footnotesize \textbf{\rotatebox{90}{ (Ours)}}}
          \put(0.5,14){\footnotesize \textbf{\rotatebox{90}{ HAG-Net \&}}}
          \put(2,14){\footnotesize \textbf{\rotatebox{90}{XGrad-CAM}}}
          \put(3.5,17){\footnotesize \textbf{\rotatebox{90}{ \cite{fu2020axiom}}}}
          \put(0.5,3){\footnotesize \textbf{\rotatebox{90}{ HAG-Net \&}}}
          \put(2,3){\footnotesize \textbf{\rotatebox{90}{Grad-CAM++}}}
          \put(3.5,6.5){\footnotesize \textbf{\rotatebox{90}{ \cite{chattopadhay2018grad}}}}
		\end{overpic}
		\caption{\textbf{Heatmaps by using different visualization approaches} including Grad-CAM++ \cite{chattopadhay2018grad}, XGrad-CAM \cite{fu2020axiom}, and Grad-CAM \cite{selvaraju2017grad} in our method.
		}
	\label{FIG:9}
\end{figure*}

\myPara{Affordance with Multiple Possibilities.} We also show the visual heatmap results to demonstrate that our model can address the challenge of multiple possibilities of affordance well. As shown in Fig. \ref{FIG:x} (a), different actions interact with different regions of same the object, and our model can localize to different regions of the object based on the action category. Fig. \ref{FIG:x} (b) shows that different interactions occur in the same region of the object, and our model can also accurately localize to the region associated with the affordance of the interaction. It demonstrates that our model can reason about human intentions through human actions and positions and well address the challenge of ambiguity in affordance.

\subsection{Ablation Studies}
\label{ablation}

\myPara{Impact of Visualization Strategy.}
In this section, we also compare two more advanced visualization methods, i.e., XGrad-CAM \cite{fu2020axiom} and Grad-CAM++ \cite{chattopadhay2018grad}, to verify the effectiveness of Grad-CAM \cite{selvaraju2017grad} used in our method for affordance grounding both subjectively and objectively.

\begin{itemize}

\item [$\bm{-}$]
\textbf{XGrad-CAM }\cite{fu2020axiom}: This paper introduces two axioms: sensitivity \cite{sundararajan2017axiomatic} and consistency \cite{montavon2018methods}, and makes XGrad-CAM satisfy these two constraints. XGrad-CAM is a visualization method with class discriminative ability to highlight the relevant regions belonging to the class, which is calculated as follows:
\begin{equation}
    w_k=\sum_i\sum_j \frac{X^{k}_{ij}}{\sum_a\sum_b X^{k}_{ab}}\odot \frac{\partial Y^c}{\partial X^{k}_{ij}}, \label{eq:no24}
\end{equation}
\begin{equation}
    H^{xgradcam}_{c}(X)=\sum_k ReLU(w_k \otimes X^k). \label{eq:no25}
\end{equation}

\item [$\bm{-}$]
\textbf{Grad-CAM++ }\cite{chattopadhay2018grad}: Grad-CAM++ is a more generalized approach based on Grad-CAM and is capable of predicting better visual interpretation of CNN models and better visual localization. Furthermore, it provides a way to measure the importance of each pixel in the feature map to the overall decision of the CNN by introducing a pixel weighting of the gradients of the output to get better visualization results, which is calculated as follows:
\begin{equation}
   \alpha^{k}_{ij}=\frac{\frac{\partial^2Y^c}{(\partial X^{k}_{ij})^2}}{2 \frac{\partial^2Y^c}{(\partial X^{k}_{ij})^2}+\sum_a \sum_b X^{k}_{ab} \odot \frac{\partial^3Y^c}{(\partial X^{k}_{ij})^3}},   \label{eq:no26}
\end{equation}
\begin{equation}
    w_k=\sum_i\sum_j \alpha^{k}_{ij} \odot ReLU(\frac{\partial Y^c}{\partial X^{k}_{ij}}), \label{eq:no27}
\end{equation}
\begin{equation}
    H^{gradcam++}_{c}(X)=\sum_{k}ReLU(w_k \otimes X^{k}).
\end{equation}

\end{itemize}

The experimental results are shown in Table \ref{Table:4}. As can be seen, using Grad-CAM in our method outperforms both of the remaining two more advanced visualization methods. In the affordance grounding task setting, it outperforms XGrad-CAM \cite{fu2020axiom} and Grad-CAM++ \cite{chattopadhay2018grad} by 13.3\% and 18.9\% in terms of NSS \cite{peters2005components} metrics on the OPRA \cite{demo2vec2018cvpr} dataset, respectively. On the EPIC \cite{Damen2018EPICKITCHENS} dataset, it exceeds XGrad-CAM by 14.6\% and surpasses Grad-CAM++ by 30.8\% in terms of NSS metrics. These results demonstrate that our method can obtain more accurate region localization. Some of the visualization results are shown in Fig. \ref{FIG:9}. The heatmaps generated by XGrad-CAM and Grad-CAM++ are large and contain a large number of irrelevant interaction regions, while our results can better focus on affordance-related regions, probably because their calculations focus more on the object as a whole, whereas our method can retain local details to better focus on affordance-related local regions of the object.

\begin{table*}[t]
\caption{\textbf{Ablation results on both OPRA \cite{demo2vec2018cvpr} and EPIC \cite{Damen2018EPICKITCHENS} datasets.} For details please refer Section \ref{ablation}.}
  \label{Table:5}
  \begin{center}
  \small
  \renewcommand{\arraystretch}{1.}
  \renewcommand{\tabcolsep}{12.1pt}
  \begin{tabular}{r|cccc|cccc}
    \hline\toprule
    \multicolumn{1}{c|}{\textbf{Dataset}} & \multicolumn{4}{c|}{\textbf{OPRA} \cite{demo2vec2018cvpr}} & \multicolumn{4}{c}{\textbf{EPIC} \cite{Damen2018EPICKITCHENS}} \\ 
    \hline
    \multicolumn{1}{c|}{Method} & \emph{$K\downarrow$} & \emph{$S\uparrow$} & \emph{$A\uparrow$} & \emph{$N\uparrow$} & \emph{$K\downarrow$} & \emph{$S\uparrow$} & \emph{$A\uparrow$} & \emph{$N\uparrow$} \\ 
    \hline
    Random (a) & $1.446$ & $0.358$ & $0.798$	& $0.877$ & $1.281$ &	$0.403$ &	 $0.772$ & $0.922$ \\
    Random (b) & $1.436$ & $0.360$ & $0.802$ & $0.900$ & $1.273$ &	$0.403$ &	 $0.779$ & $0.944$ \\
    w/o hand-aided attention (a) & $1.424$ &	$0.363$	& $0.806$ &	$0.916$ & $1.253$ & $0.404$	 & $0.789$ &	$0.960$ \\
    w/o hand-aided attention (b) & $1.420$ & $0.363$ &	$0.808$ & $0.914$ & $1.248$	 &  $0.405$ & $0.791$ & $0.974$  \\
    w/o hand-aided attention (c) & $1.416$ & $0.364$ & $0.809$ & $0.928$ & $1.235$ & $0.408$	& $0.795$ &	$0.996$   \\
    w/o select frame & $1.418$ & $0.364$ & $0.806$ & $0.934$ & $1.239$ & $0.408$ & $0.793$ & $0.990$ \\
    \hline
    Max score   & $1.437$ & $0.359$ & $0.801$ & $0.890$ & $1.261$ & $0.403$ & $0.787$ & $0.966$ \\
    Average    & $1.443$ & $0.361$ & $0.799$ & $0.885$ & $1.264$ & $0.404$ & $0.781$ & $0.968$ \\
    \hline
    w/o SEM & $1.419$ &	$0.363$	& $0.809$ & 	$0.917$ & $1.241$ & $0.407$ &	$0.797$ & $1.001$ \\
    \hline
    \rowcolor{mygray}
    Ours & $\normalsize\textbf{1.409}$ &	$\normalsize\textbf{0.365}$ & $\normalsize\textbf{0.812}$ & $\normalsize\textbf{0.948}$ & $\normalsize\textbf{1.209}$ & $\normalsize\textbf{0.414}$ & $\normalsize\textbf{0.801}$ & $\normalsize\textbf{1.045}$    \\
    \hline\bottomrule
    \end{tabular}
    \end{center}
 \end{table*}

\myPara{Effectiveness of Different Modules.}
In this section, We investigate the impact of each module in our method on affordance grounding.

\begin{itemize}
\item [$\bm{-}$]
\textbf{Random (a)}: In the preprocessing phase, a single frame is randomly chosen from the input and the process is iterated three times. We also remove the hand-aided attention during the affordance learning phase.

\item [$\bm{-}$]
\textbf{Random (b)}: In preprocessing, a single frame is randomly chosen from the input and this process is iterated three times. We keep the network structure for affordance learning as in Fig. \ref{FIG:5} (A).

\item [$\bm{-}$]
\textbf{w/o hand-aided attention (a)}: We remove the hand-aided attention in the selection network during the preprocessing process, and also remove the hand-aided attention in the HAG-Net.

\item [$\bm{-}$]
\textbf{w/o hand-aided attention (b)}: We remove the hand-aided attention in the selection network during the preprocessing process, but keep it for affordance learning as shown in Fig. \ref{FIG:5}.

\item [$\bm{-}$]
\textbf{w/o hand-aided attention (c)}: There is no change in the pre-processing stage. But during the affordance learning phase, the hand-aided attention of HAG-Net is removed.

\item[$\bm{-}$]
\textbf{w/o select frame}: We input 8 video frames into our HAG-Net without selecting keyframes.

\item [$\bm{-}$]
\textbf{Max frame}: We use the frame with the highest confidence when calculating the distillation loss.

\item [$\bm{-}$]
\textbf{Average frames}: We take the average of the input three frames when calculating the distillation loss.

\item [$\bm{-}$]
\textbf{w/o SEM}: We remove the SEM from the object branch of the affordance learning network.
\end{itemize}

\par The ablation study results are shown in Table \ref{Table:5}. From the top rows we can see that using hand-aided attention matters for affordance grounding, e.g., our model achieves relative improvements of up to 8.1\%, and 13.3\% (NSS) compared to random (a) on the OPRA \cite{demo2vec2018cvpr} and EPIC \cite{Damen2018EPICKITCHENS} datasets, respectively. It proves that the position and the action of the hand provide essential clues for the affordance of learning objects from the demonstration videos. In the setting without selecting keyframes, where we input the original video frames into the network for training, it can still achieve good results but falls behind of ours. For example, our method achieves a gain of 1.5\% NSS on OPRA and 5.6\% NSS on EPIC over the model ``w/o select frame'', respectively. These results demonstrate that the pre-processing step of selecting keyframes enables the network to learn more critical cues, leading to better results.

\par Compared to using only the frame with the highest confidence, our method achieves relative improvements of 6.5\% and 8.2\% (NSS) on OPRA and EPIC datasets, respectively. Furthermore, compared to using only the input average features, our method achieves relative improvements of 7.1\% and 8.0\% (NSS) on the two datasets, respectively. These results validate that calculating the distillation loss of the object and video branches from the highest confidence frame and the average feature of the input can better transfer the affordance cues of the video to the static objects. Compared with the model without SEM, our method achieves 3.4\% and 4.4\% (NSS) relative improvements on OPRA and EPIC, respectively. SEM can allow the object branch to focus on different parts based on appearance and action class information, thereby improving the results of affordance grounding.

\myPara{Effectiveness of Hand-aided Attention in Other Models.} We also incorporate the preprocessing strategy as well as hand-aided attention in the hotspot \cite{interaction-hotspots} model to explore the impact of hand cues for affordance grounding. The results are shown in Table \ref{Table:6}, from which we can see that levearging the human hand information can also improve the performance of the existing methods. For example, using the preprocessed frames as input to the Hotspot brings a relative improvement of 1.7\%, while adding hand-aided attention to the affordance learning process of Hotspot brings a relative 1.9\% improvement. However, the performance is still inferior to ours, which is mainly because we also leverage affordance labels to modulate the features of the target image and consider the affordance clues from hand actions and positions. Consequently, our method can better locate the human-object interactions and learn affordance-related contextual information efficiently.

\section{Conclusion and Discussion}

In this paper, we propose a novel Hand-aided Affordance Grounding Network (HAG-Net) for learning the affordance of objects from demonstration videos. By leveraging the clues offered by the position and action of the hand, we address two challenging problems in affordance grounding: (i) the same object has multiple possible interaction regions; and (ii) the same region has multiple possible interactions. Experiments on two public datasets demonstrate that our method achieves state-of-the-art results for affordance grounding.

\myPara{Weakness}
Although our HAG-Net has achieved good affordance grounding performance, there are still some limitations that should be addressed in the future work. Firstly, it is not an end-to-end solution, where a separate pre-processing stage is needed. In the future, we plan to devise an end-to-end affordance grounding method, which can optimize the selection of video frames containing affordance-related information and select hand context features in the same framework. Secondly, since the training efficiency of LSTM is low, we hope to explore the latest transformer structure in the future to improve the training efficiency as well as the performance, which has been proved to be an efficient model to handle sequential data \cite{vaswani2017attention,khan2021transformers}, \cite{girdhar2021anticipative}. 

\begin{table}[t]
\caption{\textbf{Ablation study results of exploring hand-aided attention and the preprocessing strategy} in the hotspot \cite{interaction-hotspots} model on the OPRA \cite{demo2vec2018cvpr} dataset.}
  \begin{center}
 \small
  \renewcommand{\arraystretch}{1.}
  \renewcommand{\tabcolsep}{5.pt}
  \begin{tabular}{r|cccc}
    \hline\toprule
    \multicolumn{1}{c|}{\textbf{Dataset}} & \multicolumn{4}{c}{\textbf{OPRA} \cite{demo2vec2018cvpr}} \\
    \hline
    Method & $K \downarrow$ & $S \uparrow$ & $A \uparrow$  & $N \downarrow$ \\
    \hline
    Hotspot \cite{interaction-hotspots} & $1.427$ & $0.362$ & $0.806$  & $0.907$ \\
    Select frame \& Hotspot  \cite{interaction-hotspots}  & $1.425$ & $0.364$  & $0.802$ & $0.922$ \\
    Hotspot \cite{interaction-hotspots} \& hand attention & $1.422$ & $0.363$ &  $0.806$  & $0.924$    \\
   \hline
   \rowcolor{mygray}
   Ours & $\normalsize\textbf{1.409}$ & $\normalsize\textbf{0.365}$ & $\normalsize\textbf{0.812}$ & $\normalsize\textbf{0.948}$  \\
    \hline\bottomrule
    \end{tabular}
    \end{center}
  \label{Table:6}
  \end{table}

\myPara{Potential Applications}
There may be some potential applications of our method as listed below.
\begin{itemize}
    
    \item [1)]
    Our method can provide candidate operation areas for robot grasping. The problem of vision-based robot grasping has been a hot research topic in the field of robotics \cite{du2021vision,fang2020learning,mahler2017dex}. Our approach can learn human grasping habits by observing human-object interactions, enabling the robot to mimic human actions in selecting relevant areas that can be manipulated.
    
    \item [2)]
    Our method can assist visual perception agent to comprehensively understanding of the scene \cite{DBLP:conf/cvpr/ZhuZZ15,Wang_affordanceCVPR2017}. For example, the agent need to know the semantic category of each part in a scene as well as how it interacts with people and feedbacks the environment. Moreover, it can deliver a goal-oriented understanding of the object and the environment.
    
    \item [3)]
    Visual affordance grounding can also be used in virtual reality \cite{fujinawa2017computational} applications, where the object affordance heatmap can highlight the regions where human interacts with an object, together with some virtually displayed information, e.g., affordance indicators or warning signs.
    
    \item[4)]
    In the physical domain, object affordance information can be used to improve object design and manufacture. For example, objects can be made according to human-object interaction characteristics and habits to facilitate possible human-object interactions \cite{zhao2018characterizes}.

\end{itemize}

\myPara{Future Research Directions} There are several promising directions for future research on this task.

\begin{itemize}
    \item [1)]
    \textbf{Dataset}: In the future, we will consider constructing a more extensive and richer dataset that contains videos from both first-person and third-person perspectives \cite{sigurdsson2018charades}, complex backgrounds, diverse interactions,  various object classes, and human pose annotations \cite{zhang2020towards}.
    
    \item [2)]
    \textbf{Generalization}: It is also worthwhile to further investigate how to improve the generalization ability \cite{zhu2019one,zhu2020one,zhu2020self,zhu2021self,wang2020deep} of affordance grounding methods, e.g., locating the affordance-related regions precisely among unseen objects, which is of practical importance \cite{Luo2021one}.
    
    \item [3)]
    \textbf{Compatibility}: First-person perspective videos can provide a unique viewpoint of people's interactions with objects, attention, and even intentions. In contrast, third-person videos observe human actions from an objective perspective. Such a difference poses a significant challenge for one-third-person video compatibility \cite{yu2020first,sigurdsson2018actor}. Therefore, learning better affordance grounding ability from a dataset containing both first-person and third-person videos is worth further study.
    
    \item [4)]
    \textbf{Transferability}: Although third-person videos are more accessible, in cases such as robot manipulation where the agent observes the environment and objects from the first-person perspective, it is necessary to transfer the knowledge learned from third-person videos to first-person scenarios. However, transferring affordance grounding knowledge between different perspectives \cite{li2021ego} is still under-explored and of practical meaning.
    
    \item [5)]
    \textbf{Multi-Information Assistance}: The affordance grounding task is related to many visual factors and to obtain better results in practice we need to investigate affordance grounding methods with the aid of other information, such as the introduction of texture cues \cite{zhai2019deep,zhai2020deep}, depth cues \cite{zhao2020monocular}, human-related information \cite{zhang2020towards,he2020grapy}, \etc.
    
\end{itemize}

\bibliographystyle{IEEEtran}
\bibliography{IEEEtranbib}

\begin{thebibliography}{10}
\providecommand{\url}[1]{#1}
\csname url@samestyle\endcsname
\providecommand{\newblock}{\relax}
\providecommand{\bibinfo}[2]{#2}
\providecommand{\BIBentrySTDinterwordspacing}{\spaceskip=0pt\relax}
\providecommand{\BIBentryALTinterwordstretchfactor}{4}
\providecommand{\BIBentryALTinterwordspacing}{\spaceskip=\fontdimen2\font plus
\BIBentryALTinterwordstretchfactor\fontdimen3\font minus
  \fontdimen4\font\relax}
\providecommand{\BIBforeignlanguage}[2]{{%
\expandafter\ifx\csname l@#1\endcsname\relax
\typeout{** WARNING: IEEEtran.bst: No hyphenation pattern has been}%
\typeout{** loaded for the language `#1'. Using the pattern for}%
\typeout{** the default language instead.}%
\else
\language=\csname l@#1\endcsname
\fi
#2}}
\providecommand{\BIBdecl}{\relax}
\BIBdecl

\bibitem{theaff}
J.~J. Gibson, ``The theory of affordances,'' \emph{The people, place, and space
  reader}, 1979.

\bibitem{zhang2020empowering}
J.~Zhang and D.~Tao, ``Empowering things with intelligence: A survey of the
  progress, challenges, and opportunities in artificial intelligence of
  things,'' \emph{IEEE Internet of Things Journal}, 2020.

\bibitem{DBLP:journals/corr/abs-1807-06775}
\BIBentryALTinterwordspacing
M.~Hassanin, S.~Khan, and M.~Tahtali, ``Visual affordance and function
  understanding: {A} survey,'' \emph{CoRR}, vol. abs/1807.06775, 2018.
  [Online]. Available: \url{http://arxiv.org/abs/1807.06775}
\BIBentrySTDinterwordspacing

\bibitem{DBLP:journals/cviu/KjellstromRK11}
H.~Kjellstr{\"{o}}m, J.~Romero, and D.~Kragic, ``Visual object-action
  recognition: Inferring object affordances from human demonstration,''
  \emph{Comput. Vis. Image Underst.}, vol. 115, no.~1, pp. 81--90, 2011.

\bibitem{qi2017predicting}
S.~Qi, S.~Huang, P.~Wei, and S.-C. Zhu, ``Predicting human activities using
  stochastic grammar,'' in \emph{Proceedings of the IEEE International
  Conference on Computer Vision}, 2017, pp. 1164--1172.

\bibitem{vu2014predicting}
T.-H. Vu, C.~Olsson, I.~Laptev, A.~Oliva, and J.~Sivic, ``Predicting actions
  from static scenes,'' in \emph{European Conference on Computer Vision}.\hskip
  1em plus 0.5em minus 0.4em\relax Springer, 2014, pp. 421--436.

\bibitem{yamanobe2017brief}
N.~Yamanobe, W.~Wan, I.~G. Ramirez-Alpizar, D.~Petit, T.~Tsuji, S.~Akizuki,
  M.~Hashimoto, K.~Nagata, and K.~Harada, ``A brief review of affordance in
  robotic manipulation research,'' \emph{Advanced Robotics}, vol.~31, no.
  19-20, pp. 1086--1101, 2017.

\bibitem{liu2019auto}
C.~Liu, L.-C. Chen, F.~Schroff, H.~Adam, W.~Hua, A.~L. Yuille, and L.~Fei-Fei,
  ``Auto-deeplab: Hierarchical neural architecture search for semantic image
  segmentation,'' in \emph{Proceedings of the IEEE/CVF Conference on Computer
  Vision and Pattern Recognition}, 2019, pp. 82--92.

\bibitem{DBLP:KoppulaS14}
H.~S. Koppula and A.~Saxena, ``Physically grounded spatio-temporal object
  affordances,'' in \emph{ECCV}, ser. Lecture Notes in Computer Science, vol.
  8691.\hskip 1em plus 0.5em minus 0.4em\relax Springer, 2014, pp. 831--847.

\bibitem{DBLP:conf/iros/NguyenKCT17}
A.~Nguyen, D.~Kanoulas, D.~G. Caldwell, and N.~G. Tsagarakis, ``Object-based
  affordances detection with convolutional neural networks and dense
  conditional random fields,'' in \emph{International Conference on Intelligent
  Robots and Systems,(IROS)}.\hskip 1em plus 0.5em minus 0.4em\relax IEEE,
  2017, pp. 5908--5915.

\bibitem{8460902}
T.~{Do}, A.~{Nguyen}, and I.~{Reid}, ``Affordancenet: An end-to-end deep
  learning approach for object affordance detection,'' in \emph{2018 IEEE
  International Conference on Robotics and Automation (ICRA)}, 2018, pp.
  5882--5889.

\bibitem{DBLP:conf/cvpr/SawatzkySG17}
J.~Sawatzky, A.~Srikantha, and J.~Gall, ``Weakly supervised affordance
  detection,'' in \emph{Conference on Computer Vision and Pattern Recognition,
  (CVPR)}.\hskip 1em plus 0.5em minus 0.4em\relax IEEE, 2017, pp. 5197--5206.

\bibitem{heft1989affordances}
H.~Heft, ``Affordances and the body: An intentional analysis of gibson's
  ecological approach to visual perception,'' \emph{Journal for the theory of
  social behaviour}, vol.~19, no.~1, pp. 1--30, 1989.

\bibitem{2019Object}
X.~Zhao, Y.~Cao, and Y.~Kang, ``Object affordance detection with
  relationship-aware network,'' \emph{Neural Computing and Applications}, 2019.

\bibitem{DBLP:conf/iccv/YaoMF13}
B.~Yao, J.~Ma, and F.~Li, ``Discovering object functionality,'' in
  \emph{International Conference on Computer Vision(ICCV)}.\hskip 1em plus
  0.5em minus 0.4em\relax IEEE, 2013, pp. 2512--2519.

\bibitem{Wang_affordanceCVPR2017}
X.~Wang, R.~Girdhar, and A.~Gupta, ``Binge watching: Scaling affordance
  learning from sitcoms,'' in \emph{CVPR}, 2017.

\bibitem{3d-affordance}
X.~Li, S.~Liu, K.~Kim, X.~Wang, M.-H. Yang, and J.~Kautz, ``Putting humans in a
  scene: Learning affordance in 3d indoor environments,'' in \emph{CVPR}, 2019.

\bibitem{demo2vec2018cvpr}
K.~Fang, T.-L. Wu, D.~Yang, S.~Savarese, and J.~J. Lim, ``Demo2vec: Reasoning
  object affordances from online videos,'' in \emph{The IEEE Conference on
  Computer Vision and Pattern Recognition (CVPR)}, June 2018.

\bibitem{interaction-hotspots}
T.~Nagarajan, C.~Feichtenhofer, and K.~Grauman, ``Grounded human-object
  interaction hotspots from video,'' in \emph{ICCV}, 2019.

\bibitem{Damen2018EPICKITCHENS}
D.~Damen, H.~Doughty, G.~M. Farinella, S.~Fidler, A.~Furnari, E.~Kazakos,
  D.~Moltisanti, J.~Munro, T.~Perrett, W.~Price, and M.~Wray, ``Scaling
  egocentric vision: The epic-kitchens dataset,'' in \emph{European Conference
  on Computer Vision (ECCV)}, 2018.

\bibitem{yolov3}
J.~Redmon and A.~Farhadi, ``Yolov3: An incremental improvement,'' \emph{arXiv},
  2018.

\bibitem{Liu2017learning}
Z.~Liu, J.~Li, Z.~Shen, G.~Huang, S.~Yan, and C.~Zhang, ``Learning efficient
  convolutional networks through network slimming,'' in \emph{ICCV}, 2017.

\bibitem{DBLP:conf/cvpr/LiK13}
C.~Li and K.~M. Kitani, ``Pixel-level hand detection in ego-centric videos,''
  in \emph{Conference on Computer Vision and Pattern Recognition(CVPR)}.\hskip
  1em plus 0.5em minus 0.4em\relax {IEEE} Computer Society, 2013, pp.
  3570--3577.

\bibitem{DBLP:conf/cvpr/Betancourt14}
A.~Betancourt, ``A sequential classifier for hand detection in the framework of
  egocentric vision,'' in \emph{Conference on Computer Vision and Pattern
  Recognition(CVPR)}.\hskip 1em plus 0.5em minus 0.4em\relax {IEEE} Computer
  Society, 2014, pp. 600--605.

\bibitem{DBLP:journals/cviu/BetancourtMBMRR17}
A.~Betancourt, P.~Morerio, E.~I. Barakova, L.~Marcenaro, M.~Rauterberg, and
  C.~S. Regazzoni, ``Left/right hand segmentation in egocentric videos,''
  \emph{Comput. Vis. Image Underst.}, vol. 154, pp. 73--81, 2017.

\bibitem{DBLP:conf/iccv/BambachLCY15}
S.~Bambach, S.~Lee, D.~J. Crandall, and C.~Yu, ``Lending {A} hand: Detecting
  hands and recognizing activities in complex egocentric interactions,'' in
  \emph{International Conference on Computer Vision, (ICCV)}.\hskip 1em plus
  0.5em minus 0.4em\relax {IEEE} Computer Society, 2015, pp. 1949--1957.

\bibitem{9060114}
G.~{Kapidis}, R.~{Poppe}, E.~{Van Dam}, L.~{Noldus}, and R.~{Veltkamp},
  ``Egocentric hand track and object-based human action recognition,'' in
  \emph{2019 IEEE SmartWorld, Ubiquitous Intelligence Computing, Advanced
  Trusted Computing, Scalable Computing Communications, Cloud Big Data
  Computing, Internet of People and Smart City Innovation
  (SmartWorld/SCALCOM/UIC/ATC/CBDCom/IOP/SCI)}, 2019, pp. 922--929.

\bibitem{DBLP:conf/icvs/StarkLZWS08}
M.~Stark, P.~Lies, M.~Zillich, J.~L. Wyatt, and B.~Schiele, ``Functional object
  class detection based on learned affordance cues,'' in \emph{Computer Vision
  Systems, 6th International Conference(ICVS)}, 2008, pp. 435--444.

\bibitem{DBLP:journals/ras/0004RL14}
Y.~Sun, S.~Ren, and Y.~Lin, ``Object-object interaction affordance learning,''
  \emph{Robotics Auton. Syst.}, vol.~62, no.~4, pp. 487--496, 2014.

\bibitem{DBLP:conf/icra/SongKOPABK13}
D.~Song, N.~Kyriazis, I.~Oikonomidis, C.~Papazov, A.~A. Argyros, D.~Burschka,
  and D.~Kragic, ``Predicting human intention in visual observations of
  hand/object interactions,'' in \emph{International Conference on Robotics and
  Automation(ICRA)}.\hskip 1em plus 0.5em minus 0.4em\relax IEEE, 2013, pp.
  1608--1615.

\bibitem{DBLP:conf/icra/PieropanEK13}
A.~Pieropan, C.~H. Ek, and H.~Kjellstr{\"{o}}m, ``Functional object descriptors
  for human activity modeling,'' in \emph{International Conference on Robotics
  and Automation(ICRA)}.\hskip 1em plus 0.5em minus 0.4em\relax IEEE, 2013, pp.
  1282--1289.

\bibitem{schulman2016learning}
J.~Schulman, J.~Ho, C.~Lee, and P.~Abbeel, ``Learning from demonstrations
  through the use of non-rigid registration,'' in \emph{Robotics
  Research}.\hskip 1em plus 0.5em minus 0.4em\relax Springer, 2016, pp.
  339--354.

\bibitem{chu2016learning}
V.~Chu, B.~Akgun, and A.~L. Thomaz, ``Learning haptic affordances from
  demonstration and human-guided exploration,'' in \emph{2016 IEEE haptics
  symposium (HAPTICS)}.\hskip 1em plus 0.5em minus 0.4em\relax IEEE, 2016, pp.
  119--125.

\bibitem{aleotti2011part}
J.~Aleotti and S.~Caselli, ``Part-based robot grasp planning from human
  demonstration,'' in \emph{2011 IEEE International Conference on Robotics and
  Automation}.\hskip 1em plus 0.5em minus 0.4em\relax IEEE, 2011, pp.
  4554--4560.

\bibitem{zha2021contrastively}
Y.~Zha, S.~Bhambri, and L.~Guan, ``Contrastively learning visual attention as
  affordance cues from demonstrations for robotic grasping,'' \emph{arXiv
  preprint arXiv:2104.00878}, 2021.

\bibitem{chen2020recursive}
Z.~Chen, J.~Zhang, and D.~Tao, ``Recursive context routing for object
  detection,'' \emph{International Journal of Computer Vision}, pp. 1--19,
  2020.

\bibitem{ren2016faster}
S.~Ren, K.~He, R.~Girshick, and J.~Sun, ``Faster r-cnn: towards real-time
  object detection with region proposal networks,'' \emph{IEEE transactions on
  pattern analysis and machine intelligence}, vol.~39, no.~6, pp. 1137--1149,
  2016.

\bibitem{DBLP:conf/bmvc/MittalZT11}
A.~Mittal, A.~Zisserman, and P.~H.~S. Torr, ``Hand detection using multiple
  proposals,'' in \emph{British Machine Vision Conference, (BMVC)}, J.~Hoey,
  S.~J. McKenna, and E.~Trucco, Eds., 2011, pp. 1--11.

\bibitem{DBLP:conf/cvpr/HeZRS16}
K.~He, X.~Zhang, S.~Ren, and J.~Sun, ``Deep residual learning for image
  recognition,'' in \emph{(IEEE) Conference on Computer Vision and Pattern
  Recognition, (CVPR)}, 2016, pp. 770--778.

\bibitem{hochreiter1997long}
S.~Hochreiter and J.~Schmidhuber, ``Long short-term memory,'' \emph{Neural
  computation}, vol.~9, no.~8, pp. 1735--1780, 1997.

\bibitem{singh2019hetconv}
P.~Singh, V.~K. Verma, P.~Rai, and V.~P. Namboodiri, ``Hetconv: Heterogeneous
  kernel-based convolutions for deep cnns,'' in \emph{Proceedings of the
  IEEE/CVF Conference on Computer Vision and Pattern Recognition}, 2019, pp.
  4835--4844.

\bibitem{sun2019high}
K.~Sun, Y.~Zhao, B.~Jiang, T.~Cheng, B.~Xiao, D.~Liu, Y.~Mu, X.~Wang, W.~Liu,
  and J.~Wang, ``High-resolution representations for labeling pixels and
  regions,'' \emph{arXiv preprint arXiv:1904.04514}, 2019.

\bibitem{wu2018group}
Y.~Wu and K.~He, ``Group normalization,'' in \emph{Proceedings of the European
  conference on computer vision (ECCV)}, 2018, pp. 3--19.

\bibitem{selvaraju2017grad}
R.~R. Selvaraju, M.~Cogswell, A.~Das, R.~Vedantam, D.~Parikh, and D.~Batra,
  ``Grad-cam: Visual explanations from deep networks via gradient-based
  localization,'' in \emph{Proceedings of the IEEE international conference on
  computer vision}, 2017, pp. 618--626.

\bibitem{bylinskii2018different}
Z.~Bylinskii, T.~Judd, A.~Oliva, A.~Torralba, and F.~Durand, ``What do
  different evaluation metrics tell us about saliency models?'' \emph{IEEE
  transactions on pattern analysis and machine intelligence}, vol.~41, no.~3,
  pp. 740--757, 2018.

\bibitem{swain1991color}
M.~J. Swain and D.~H. Ballard, ``Color indexing,'' \emph{International journal
  of computer vision}, vol.~7, no.~1, pp. 11--32, 1991.

\bibitem{DBLP:conf/iccv/JuddEDT09}
T.~Judd, K.~A. Ehinger, F.~Durand, and A.~Torralba, ``Learning to predict where
  humans look,'' in \emph{ICCV}.\hskip 1em plus 0.5em minus 0.4em\relax {IEEE}
  Computer Society, 2009, pp. 2106--2113.

\bibitem{peters2005components}
R.~J. Peters, A.~Iyer, L.~Itti, and C.~Koch, ``Components of bottom-up gaze
  allocation in natural images,'' \emph{Vision research}, vol.~45, no.~18, pp.
  2397--2416, 2005.

\bibitem{DBLP:conf/eccv/HuangCLS18}
Y.~Huang, M.~Cai, Z.~Li, and Y.~Sato, ``Predicting gaze in egocentric video by
  learning task-dependent attention transition,'' in \emph{European Conference
  on Computer Vision (ECCV)}, 2018.

\bibitem{DBLP:conf/icpr/CorniaBSC16}
M.~Cornia, L.~Baraldi, G.~Serra, and R.~Cucchiara, ``A deep multi-level network
  for saliency prediction,'' in \emph{ICPR}, 2016.

\bibitem{DBLP:journals/corr/KummererWB16}
M.~K{\"{u}}mmerer, T.~S.~A. Wallis, and M.~Bethge, ``Deepgaze {II:} reading
  fixations from deep features trained on object recognition,'' \emph{CoRR},
  vol. abs/1610.01563, 2016.

\bibitem{Pan_2017_SalGAN}
J.~Pan, C.~Canton, K.~McGuinness, N.~E. O'Connor, J.~Torres, E.~Sayrol, and
  X.~a. Giro-i Nieto, ``Salgan: Visual saliency prediction with generative
  adversarial networks,'' in \emph{arXiv}, January 2017.

\bibitem{DBLP:journals/corr/SimonyanZ14a}
K.~Simonyan and A.~Zisserman, ``Very deep convolutional networks for
  large-scale image recognition,'' in \emph{ICRL}, 2015.

\bibitem{DBLP:conf/cvpr/DengDSLL009}
J.~Deng, W.~Dong, R.~Socher, L.~Li, K.~Li, and F.~Li, ``Imagenet: {A}
  large-scale hierarchical image database,'' in \emph{Computer Society
  Conference on Computer Vision and Pattern Recognition (CVPR)}.\hskip 1em plus
  0.5em minus 0.4em\relax {IEEE} Computer Society, 2009, pp. 248--255.

\bibitem{chattopadhay2018grad}
A.~Chattopadhay, A.~Sarkar, P.~Howlader, and V.~N. Balasubramanian,
  ``Grad-cam++: Generalized gradient-based visual explanations for deep
  convolutional networks,'' in \emph{2018 IEEE Winter Conference on
  Applications of Computer Vision (WACV)}.\hskip 1em plus 0.5em minus
  0.4em\relax IEEE, 2018, pp. 839--847.

\bibitem{fu2020axiom}
R.~Fu, Q.~Hu, X.~Dong, Y.~Guo, Y.~Gao, and B.~Li, ``Axiom-based grad-cam:
  Towards accurate visualization and explanation of cnns,'' \emph{arXiv
  preprint arXiv:2008.02312}, 2020.

\bibitem{sundararajan2017axiomatic}
M.~Sundararajan, A.~Taly, and Q.~Yan, ``Axiomatic attribution for deep
  networks,'' in \emph{International Conference on Machine Learning}.\hskip 1em
  plus 0.5em minus 0.4em\relax PMLR, 2017, pp. 3319--3328.

\bibitem{montavon2018methods}
G.~Montavon, W.~Samek, and K.-R. M{\"u}ller, ``Methods for interpreting and
  understanding deep neural networks,'' \emph{Digital Signal Processing},
  vol.~73, pp. 1--15, 2018.

\bibitem{vaswani2017attention}
A.~Vaswani, N.~Shazeer, N.~Parmar, J.~Uszkoreit, L.~Jones, A.~N. Gomez,
  L.~Kaiser, and I.~Polosukhin, ``Attention is all you need,'' \emph{arXiv
  preprint arXiv:1706.03762}, 2017.

\bibitem{khan2021transformers}
S.~Khan, M.~Naseer, M.~Hayat, S.~W. Zamir, F.~S. Khan, and M.~Shah,
  ``Transformers in vision: A survey,'' \emph{arXiv preprint arXiv:2101.01169},
  2021.

\bibitem{girdhar2021anticipative}
R.~Girdhar and K.~Grauman, ``Anticipative video transformer,'' \emph{arXiv
  preprint arXiv:2106.02036}, 2021.

\bibitem{du2021vision}
G.~Du, K.~Wang, S.~Lian, and K.~Zhao, ``Vision-based robotic grasping from
  object localization, object pose estimation to grasp estimation for parallel
  grippers: a review,'' \emph{Artificial Intelligence Review}, vol.~54, no.~3,
  pp. 1677--1734, 2021.

\bibitem{fang2020learning}
K.~Fang, Y.~Zhu, A.~Garg, A.~Kurenkov, V.~Mehta, L.~Fei-Fei, and S.~Savarese,
  ``Learning task-oriented grasping for tool manipulation from simulated
  self-supervision,'' \emph{The International Journal of Robotics Research},
  vol.~39, no. 2-3, pp. 202--216, 2020.

\bibitem{mahler2017dex}
J.~Mahler, J.~Liang, S.~Niyaz, M.~Laskey, R.~Doan, X.~Liu, J.~A. Ojea, and
  K.~Goldberg, ``Dex-net 2.0: Deep learning to plan robust grasps with
  synthetic point clouds and analytic grasp metrics,'' \emph{arXiv preprint
  arXiv:1703.09312}, 2017.

\bibitem{DBLP:conf/cvpr/ZhuZZ15}
Y.~Zhu, Y.~Zhao, and S.~Zhu, ``Understanding tools: Task-oriented object
  modeling, learning and recognition,'' in \emph{Conference on Computer Vision
  and Pattern Recognition, (CVPR)}.\hskip 1em plus 0.5em minus 0.4em\relax
  {IEEE}, 2015, pp. 2855--2864.

\bibitem{fujinawa2017computational}
E.~Fujinawa, S.~Yoshida, Y.~Koyama, T.~Narumi, T.~Tanikawa, and M.~Hirose,
  ``Computational design of hand-held vr controllers using haptic shape
  illusion,'' in \emph{Proceedings of the 23rd acm symposium on virtual reality
  software and technology}, 2017, pp. 1--10.

\bibitem{zhao2018characterizes}
N.~Zhao, Y.~Cao, and R.~W. Lau, ``What characterizes personalities of graphic
  designs?'' \emph{ACM Transactions on Graphics (TOG)}, vol.~37, no.~4, pp.
  1--15, 2018.

\bibitem{sigurdsson2018charades}
G.~A. Sigurdsson, A.~Gupta, C.~Schmid, A.~Farhadi, and K.~Alahari,
  ``Charades-ego: A large-scale dataset of paired third and first person
  videos,'' \emph{arXiv preprint arXiv:1804.09626}, 2018.

\bibitem{zhang2020towards}
J.~Zhang, Z.~Chen, and D.~Tao, ``Towards high performance human keypoint
  detection,'' \emph{International Journal of Computer Vision}, pp. 1--24,
  2021.

\bibitem{zhu2019one}
K.~Zhu, W.~Zhai, Z.-J. Zha, and Y.~Cao, ``One-shot texture retrieval with
  global context metric,'' \emph{IJCAI}, 2019.

\bibitem{zhu2020one}
K.~Zhu, Y.~Cao, W.~Zhai, and Z.-J. Zha, ``One-shot texture retrieval using
  global grouping metric,'' \emph{IEEE Transactions on Multimedia}, 2020.

\bibitem{zhu2020self}
K.~Zhu, W.~Zhai, and Y.~Cao, ``Self-supervised tuning for few-shot
  segmentation,'' in \emph{Twenty-Ninth International Joint Conference on
  Artificial Intelligence {IJCAI-20}}, 2020.

\bibitem{zhu2021self}
K.~Zhu, Y.~Cao, W.~Zhai, J.~Cheng, and Z.-J. Zha, ``Self-promoted prototype
  refinement for few-shot class-incremental learning,'' in \emph{Proceedings of
  the IEEE/CVF Conference on Computer Vision and Pattern Recognition}, 2021,
  pp. 6801--6810.

\bibitem{wang2020deep}
W.~Wang, W.~Zhai, and Y.~Cao, ``Deep inhomogeneous regularization for transfer
  learning,'' in \emph{2020 IEEE International Conference on Image Processing
  (ICIP)}.\hskip 1em plus 0.5em minus 0.4em\relax IEEE, 2020, pp. 221--225.

\bibitem{Luo2021one}
H.~Luo, W.~Zhai, J.~Zhang, Y.~Cao, and D.~Tao, ``One-shot affordance
  detection,'' \emph{IJCAI}, 2021.

\bibitem{yu2020first}
H.~Yu, M.~Cai, Y.~Liu, and F.~Lu, ``First-and third-person video co-analysis by
  learning spatial-temporal joint attention,'' \emph{IEEE Transactions on
  Pattern Analysis and Machine Intelligence}, 2020.

\bibitem{sigurdsson2018actor}
G.~A. Sigurdsson, A.~Gupta, C.~Schmid, A.~Farhadi, and K.~Alahari, ``Actor and
  observer: Joint modeling of first and third-person videos,'' in
  \emph{Proceedings of the IEEE Conference on Computer Vision and Pattern
  Recognition}, 2018, pp. 7396--7404.

\bibitem{li2021ego}
Y.~Li, T.~Nagarajan, B.~Xiong, and K.~Grauman, ``Ego-exo: Transferring visual
  representations from third-person to first-person videos,'' \emph{arXiv
  preprint arXiv:2104.07905}, 2021.

\bibitem{zhai2019deep}
W.~Zhai, Y.~Cao, J.~Zhang, and Z.-J. Zha, ``Deep multiple-attribute-perceived
  network for real-world texture recognition,'' in \emph{Proceedings of the
  IEEE/CVF International Conference on Computer Vision}, 2019, pp. 3613--3622.

\bibitem{zhai2020deep}
W.~Zhai, Y.~Cao, Z.-J. Zha, H.~Xie, and F.~Wu, ``Deep structure-revealed
  network for texture recognition,'' in \emph{Proceedings of the IEEE/CVF
  Conference on Computer Vision and Pattern Recognition}, 2020, pp.
  11\,010--11\,019.

\bibitem{zhao2020monocular}
C.~Zhao, Q.~Sun, C.~Zhang, Y.~Tang, and F.~Qian, ``Monocular depth estimation
  based on deep learning: An overview,'' \emph{Science China Technological
  Sciences}, pp. 1--16, 2020.

\bibitem{he2020grapy}
H.~He, J.~Zhang, Q.~Zhang, and D.~Tao, ``Grapy-ml: graph pyramid mutual
  learning for cross-dataset human parsing,'' in \emph{Proceedings of the AAAI
  Conference on Artificial Intelligence}, vol.~34, no.~07, 2020, pp.
  10\,949--10\,956.

\end{thebibliography}

\end{document}